\documentclass{article}

     \PassOptionsToPackage{numbers, compress}{natbib}

\usepackage[preprint]{neurips_data_2024}
\usepackage{./macros/packages}
\usepackage{./macros/editing-macros}
\usepackage{./macros/formatting}
\usepackage{./macros/statistics-macros}
\usepackage{multirow}

\usepackage{pifont}
\newcommand{\cmark}{\ding{51}}
\newcommand{\xmark}{\ding{55}}

\usepackage[textsize=tiny]{todonotes}

\usepackage[utf8]{inputenc} 
\usepackage[T1]{fontenc}    
\usepackage{hyperref}       
\usepackage{url}            
\usepackage{booktabs}       
\usepackage{amsfonts}       
\usepackage{nicefrac}       
\usepackage{microtype}      
\usepackage{xcolor}         

\usepackage{color}
\usepackage{listings}
\DeclareFixedFont{\ttb}{T1}{txtt}{bx}{n}{9.5} 
\DeclareFixedFont{\ttm}{T1}{txtt}{m}{n}{9.5}  
\definecolor{codeblue}{rgb}{0,0,0.6}
\definecolor{codegreen}{rgb}{0,0.6,0}
\definecolor{dark-blue}{rgb}{0.15,0.15,0.4}
\definecolor{codepurple}{rgb}{0.6,0,0.6}

\newcommand\pythonstyle{\lstset{
    language=Python,
    basicstyle=\scriptsize\ttfamily,
    otherkeywords={self,with},             
    keywordstyle=\color{codepurple},
    emph={__init__, dim, None},
    emphstyle=\color{codeblue},
    stringstyle=\color{codegreen},
    commentstyle=\color{codegreen},
    frame=none,              
    showstringspaces=false,
    breaklines=true,
    numbers=left,
    numbersep=3pt,
    tabsize=2,
    breakatwhitespace=false,
    abovecaptionskip=2ex,
    captionpos=b,
}}

\lstnewenvironment{python}[1][]
{
    
    \pythonstyle
    \lstset{#1}
}{}

\title{AExGym: Benchmarks and Environments  \\ for Adaptive Experimentation}

\author{%
    Jimmy Wang \\
Department of Computer Science\\
Columbia University\\
\texttt{jw4209@columbia.edu} \\
\And
Ethan Che \\
Decision, Risk, and Operations Division\\
Columbia Business School\\
\texttt{ewc2119@columbia.edu} \\
\And
Daniel R. Jiang \\
Meta, University of Pittsburgh\\
\texttt{drjiang@meta.com}
\And
Hongseok Namkoong \\
Decision, Risk, and Operations Division\\
Columbia Business School\\
\texttt{namkoong@gsb.columbia.edu} 
}

\begin{document}

\maketitle

\begin{abstract}

Innovations across science and industry are evaluated using randomized trials (i.e., A/B tests). While simple and robust, such static designs are inefficient or infeasible for testing many hypotheses. Adaptive designs can greatly improve statistical power in theory, but they have seen limited adoption due to their fragility in practice. We present a benchmark for adaptive experimentation based on real-world datasets, highlighting prominent practical challenges to operationalizing adaptivity: non-stationarity, batched/delayed feedback, multiple outcomes and objectives, and external validity. Our benchmark aims to spur methodological development that puts practical performance (e.g., robustness) as a central concern, rather than mathematical guarantees on contrived instances. We release an open-source library, AExGym, which is designed with modularity and extensibility in mind to allow experimentation practitioners to develop and benchmark custom environments and algorithms.


\end{abstract}

\section{Introduction}

Experimentation is the foundation of scientific discovery. Standard experiments utilize static designs (also known as a randomized control trial or an A/B test) where units are randomly assigned to treatment arms proportionate to a fixed probability. Due to resource constraints and statistical power considerations~\cite{ioannidis2017power}, static designs typically only allow comparing between one to two treatment options. The ability to simultaneously test many different treatment arms can significantly accelerate innovation, yet the cost of experimentation is often prohibitive. In social sciences, 
``the list of interventions under consideration is so long that it is prohibitively costly and time consuming to sufficiently test the full range of treatment arms'' \cite{OfferWestortCoGr20}. 
Even in the tech industry where it is commonplace to experiment on millions of units, power considerations allow testing only a handful of treatment options \citep{KohaviLoSoHe09,gupta2019top,kohavi2020trustworthy,liu2021datasets}.




Adaptive experimentation can dramatically improve statistical power by dynamically allocating more resources to promising arms as more data is gathered~\cite{GlynnJu04, JamiesonNo14, Russo20}.
When it is possible to change the allocation many times (``long horizon''), the vast \emph{multi-armed bandits} (MABs) literature develops adaptive designs that enjoy strong mathematical guarantees in idealistic settings~\cite{HongNeXu15, ChenChLePu15, LattimoreSz19}. 
In particular, when sample sizes are small compared to the population and within-experimental regret is less of a priority, \emph{best-arm identification} (BAI) methods enjoy strong optimality guarantees for identifying the best arm at the end of the experiment \cite{MannorTs04,EvenDarMaMa06,BubeckMuSt09, AudibertBuMu10,GabillonGhLa12,
  KarninKoSo13, JamiesonMaNoBu14, JamiesonNo14, KaufmannKa13, GarivierKa16,
  KaufmannCaGa16,
CarpentierLo16,Russo20}. 

However, the fully interactive methods in the MAB literature have found limited practical impact in adaptive experimentation where there is only a handful of opportunities to change the allocation.
Static designs continue to be the dominant type of experiments in real-world settings. 
Existing experimentation systems largely support standard A/B testing; large technology companies often have centralized experimentation platforms that are used by the entire company, such as LinkedIn XLNT \citep{xu2015infrastructure}, Microsoft ExP \citep{gupta2018anatomy}, or Netflix XP \citep{diamantopoulos2020engineering}. Despite well-known applications of adaptive experimentation (e.g., Meta \citep{bakshy2018ae,bakshy2019ax}, Amazon \citep{fiez2024best}), it is anecdotally understood that experimenters are often hesitant due to a fear of potential downsides \citep{OfferWestortCoGr20,fiez2024best} and in general, a lack of confidence in their effectiveness outside of idealized environments.

\begin{figure}
    \centering
    \includegraphics[width=0.99\textwidth]{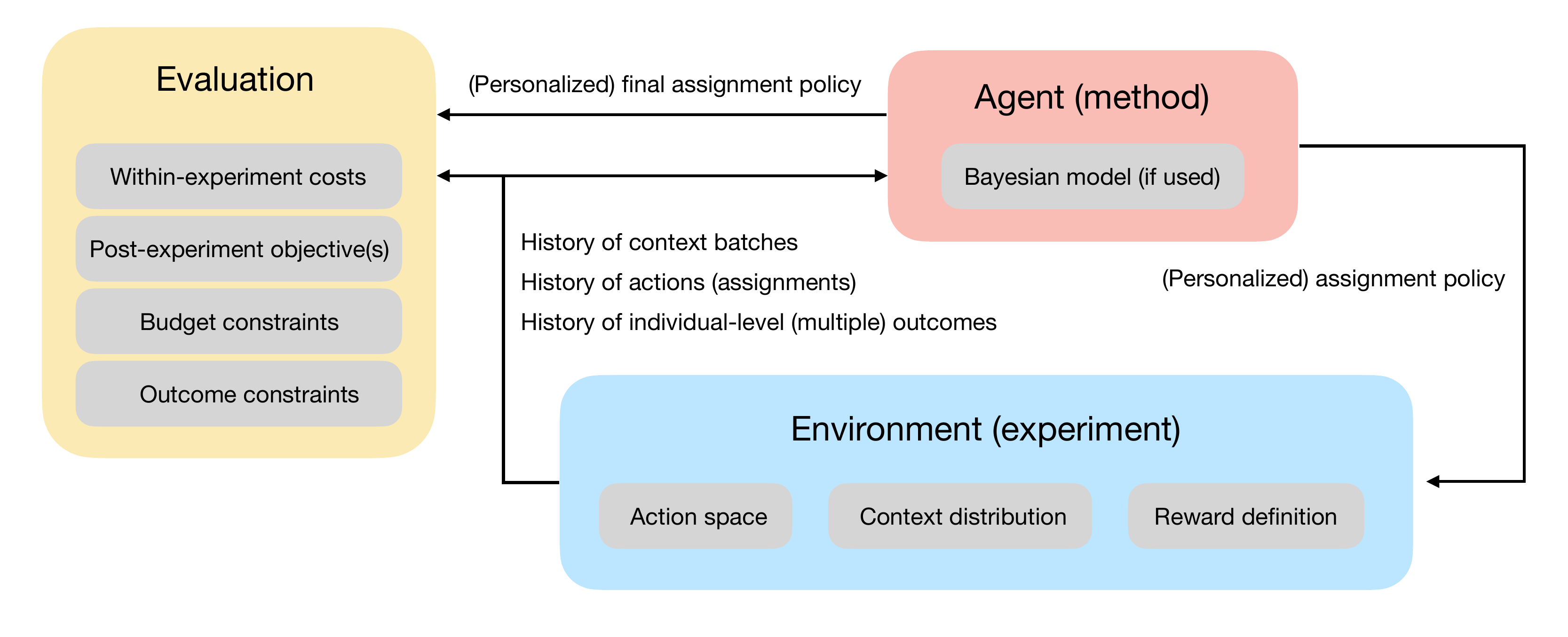}
    \vspace{-6pt}
    \caption{The design of $\mathsf{AExGym}$ includes an $\mathsf{Environment}$, $\mathsf{Agent}$, and a set of criteria for evaluation. In each period, the $\mathsf{Environment}$ generates a batch of contexts (e.g., user features). The $\mathsf{Agent}$ receives these contexts, along with the entire history of the experimentation process (past context batches, assignments, and individual-level outcomes), and outputs a possibly personalized assignment policy. At the end of the experiment, the history of the experimentation process together with a final assignment policy from the $\mathsf{Agent}$ are evaluated by a set of practical evaluation criteria that may include one or more of the following: within-experiment costs, post-experiment objectives, budget constraints, or outcome constraints.}
    \label{fig:interface}
    \vspace{-12pt}
\end{figure}

As we demonstrate in the sequel, the practitioners' reluctance is well-founded. Real-world experiments are  not conducted in idealized environments required for standard theoretical guarantees \citep{LattimoreSz19}.
Outcomes are measured in large batches (or ``waves'') due to delayed feedback and the practical cost of updating treatment allocations. 
Environments suffer from significant, naturally occurring nonstationarity (e.g., day-of-the-week effects) where traditional adaptive algorithms fail~\cite{QinRusso2023, fiez2024best}. 
Instead of the classical dichotomy between 
within- and post-experiment regret over a single outcome, practitioners need to balance these goals defined over multiple outcomes under various constraints, such as collecting sufficient samples for post-experiment inference \citep{OfferWestortCoGr20, ZhangJaMu20}. In addition, when assessing different treatments, experimenters rarely care about a single outcome variable and rather must balance trade-offs between several outcomes simultaneously~\cite{daulton20diff, daulton21par}.
Finally, in addition to internal validity over the experimental population picked out of convenience, researchers want \emph{external validity}, the ability to generalize findings to the broader population. 

There is a growing body of work that addresses the aforementioned practical challenges, but the standard approach to methodological development focusing on theoretical bounds is brittle. 
Under this \emph{deductive approach}, even small variations to the problem setting 
requires highly trained methodological researchers to make ad-hoc adjustments to algorithms.
As an example, several authors recently develop specialized algorithms to address  individual issues (e.g.,
batching~\cite{SculleyEtAl15, AgarwalEtAl16,
  BakshyEtAl18, NamkoongDaBa20}, nonstationarity~\cite{QinRusso2023}),
yet we are unaware of an algorithm that can address even two of the aforementioned challenges.
In contrast, the standard scientific methodology in ML is \emph{inductive} in nature. Algorithms are developed based on empirical benchmarking, 
where methological innovations are generated based on observations from real datasets.

To promote inductive investigations that model practical challenges, we must have corresponding empirical foundations.
This paper introduces the $\mathsf{AExGym}$ framework, an open-source software package meant to allow experimenters to easily evaluate algorithms on a range of real-world settings and develop new algorithmic approaches that can handle a myriad of practical challenges. In particular, our package offers a 
initial suite of environments that model the most salient challenges we identified:
batched feedback, nonstationarity,  multiple objectives and outcomes, constraints, and external validity.

\begin{table}
\small
\centering
\begin{tabular}{ccccc}
\toprule 
 & $\begin{array}{c}
\text{Vowpal } 
\text{Wabbit \citep{vowpal2012}}
\end{array}$ & $\begin{array}{c}
\text{Open Bandit \citep{saito2021open}}
\end{array}$ & RecoGym \citep{rohde2018recogym} & Ours\tabularnewline
\midrule
\midrule 
Offline data & \textcolor{green}{$\text{{\cmark}}$} & \textcolor{green}{$\text{{\cmark}}$} & \textcolor{red}{$\text{{\xmark}}$} & \textcolor{green}{$\text{{\cmark}}$}\tabularnewline
\midrule 
Multiple outcomes & \textcolor{red}{$\text{{\xmark}}$} & \textcolor{red}{$\text{{\xmark}}$} & \textcolor{red}{$\text{{\xmark}}$} & \textcolor{green}{$\text{{\cmark}}$}\tabularnewline
\midrule 
Non-stationarity & \textcolor{red}{$\text{{\xmark}}$} & \textcolor{red}{$\text{{\xmark}}$} & \textcolor{red}{$\text{{\xmark}}$} & \textcolor{green}{$\text{{\cmark}}$}\tabularnewline
\midrule 
Batched feedback & \textcolor{green}{$\text{{\cmark}}$} & \textcolor{green}{$\text{{\cmark}}$} & \textcolor{green}{$\text{{\cmark}}$} & \textcolor{green}{$\text{{\cmark}}$}\tabularnewline
\midrule 
Custom objectives & \textcolor{red}{$\text{{\xmark}}$} & \textcolor{red}{$\text{{\xmark}}$} & \textcolor{red}{$\text{{\xmark}}$} & \textcolor{green}{$\text{{\cmark}}$}\tabularnewline
\midrule 
A/B test to MAB pipeline & \textcolor{red}{$\text{{\xmark}}$} & \textcolor{red}{$\text{{\xmark}}$} & \textcolor{red}{$\text{{\xmark}}$} & \textcolor{green}{$\text{{\cmark}}$}\tabularnewline
\midrule 
Best-arm identification & \textcolor{red}{$\text{{\xmark}}$} & \textcolor{red}{$\text{{\xmark}}$} & \textcolor{red}{$\text{{\xmark}}$} & \textcolor{green}{$\text{{\cmark}}$}\tabularnewline
\bottomrule
\end{tabular}
\vspace{1em}
\caption{Comparison of $\mathsf{AExGym}$ with existing bandit environments for features relevant for adaptive experimentation. {\bf Offline data} indicates compatibility with offline data. {\bf Multiple metrics} indicates support for multiple metrics/rewards. {\bf Non-stationarity} indicates functionality for non-stationary environments and automatic construction of temporal features for contextual policies. {\bf Batched feedback} refers to batched updates to the policy. {\bf Custom objectives} indicate support for custom experimental objectives such as cumulative regret, simple regret, policy regret, top-K arm selection, etc. {\bf A/B test to MAB pipeline} refers to functionality for converting A/B test data (with a single treatment and control) to a multi-armed bandit instance. {\bf Best-arm identification} indicates implementations of BAI policies.}
\label{table:comparison}
\end{table}
Our primary observation is that adaptive experiments can be formulated using the language of Markov decision processes \citep{frazier2008knowledge,lam2016bayesian,min2019thompson,jiang2020efficient,min2020policy,che2023adaptive}. We design a standardized interface for the adaptive experimentation process, and implement APIs for testing across a range  \emph{environments} that may arise in practice.
Instead of operating on a typical ``policy'' as is standard in reinforcement learning,  we use the notion of an \emph{agent} to model \emph{adaptive experimentation algorithms} that update policies as it gathers more data. Moving away from the traditional focus on a singular notion of performance (e.g., within- vs. post-experiment), our framework allows flexible \emph{evaluation criteria} which enables a self-contained and comprehensive benchmark. See Figure \ref{fig:interface} for an illustration of how these pieces fit together and Section \ref{sec:aexgym} for a detailed description.
Our framework is inspired by and bears high-level resemblance to OpenAI Gym ~\cite{BrockmanChPeScScTaZa16} which provides standard APIs for reinforcement learning (RL).

$\mathsf{AExGym}$ provides a user-friendly interface to construct an environment from offline data. To demonstrate this, we construct benchmarks using real-world, publicly-available data across a range of application domains. Specifically, we model challenges involving geographical constraints, external validity, and personalization by repurposing data from a multi-site study of microcredit expansions \cite{Meager2019Understanding}, the Pennsylvania Reemployment Bonus Demonstration \cite{corson1991pennsylvania}, and the National Health Interview Survey \cite{NHIS2010-2019}. Furthermore, we curate experimental data involving naturally occurring non-stationarity and realistic treatment effect sizes, by adapting over 240 settings in the ASOS.com dataset \cite{liu2021datasets} and gathering statistics from 21 field experiments ranging from education to welfare \cite{wahlstrom2014examining, weiss2017effects, Riccio1992GAIN, Hamilton2001National, Kemple1995Florida, Ariel2016Wearing}. Using these settings, we construct realistic examples involving batching, multiple objectives, and constraints.  

These benchmarks are designed to develop an understanding of when adaptive algorithms can meaningfully improve upon static designs 
in realistic experimental settings. 
We hope to spur inductive approaches to algorithmic development that address the myriad of challenges that arise in practice, and that our benchmark can provide experimenters who are considering deploying adaptive designs with solid empirical evidence of their performance.

\textbf{Related Work.}
Existing benchmarks and datasets for adaptive algorithms primarily focus on the idealized, fully-interactive multi-armed bandit setting, where the goal is to minimize within-experiment (cumulative) regret over a long horizon. 
For example, there are numerous large-scale datasets for benchmarking bandit algorithms for recommendation systems~\citep{li2010contextual, lefortier2016large, wu2020mind, saito2021open, gao2022kuairec}. Other related benchmarks convert traditional supervised learning datasets into an online learning problem \cite{bietti2021bakeoff, riquelme2018deep}; while these works allow generating insights on a broad set of problem settings, fictitious settings constructed from supervised datasets do not model practical challenges and are often considered simplistic and easy instances.

In contrast to the fully interactive setting,
this work focuses on improving static designs by injecting a few rounds of adaptivity. 
Our framework is thus geared toward adaptive experiments in the A/B testing domain, where practitioners may have only a few opportunities to reallocate samples due to operational costs. 
To our knowledge, we provide the first complete benchmark and a corresponding software framework tailored to adaptive experimentation.
The set of initial environments we provide in our benchmark is built on real-world experiments with natural treatment effect sizes and practical challenges faced by experimentation practitioners.

There are also a number of related software packages that implement adaptive algorithms, such as those that specialize in contextual bandits \citep{vowpal2012,saito2021open}, Bayesian optimization \citep{balandat2020botorch,dragonfly2020}, and reinforcement learning \citep{rohde2018recogym,liang2018rllib,raffin2021stable,huang2022cleanrl,weng2022tianshou,bou2023torchrl}. A notable package is Ax \citep{bakshy2018ae,bakshy2019ax}, which is an industry-grade platform that provides the infrastructure (i.e., data management, deployment, storage) for adaptive experiments. It largely takes a ``parameter-tuning view'' of adaptive experimentation (it uses BoTorch \citep{balandat2020botorch} under-the-hood), where an expensive-to-evaluate function $f(x)$ is being optimized over a continuous space. This differs from our view of adaptive experimentation, where we optimize over policies that perform individual-level assignments to a finite number of treatments. Another package is AEPsych~\cite{owen21psycho}, an experimentation platform for human perception experiments built on top of Ax.



Table \ref{table:comparison} compares $\mathsf{AExGym}$ to three package that are most similar to it in terms of features: Vowpal Wabbit \citep{vowpal2012}, Open Bandit Dataset/Pipeline \citep{saito2021open}, and RecoGym \citep{rohde2018recogym}.
Our package $\mathsf{AExGym}$ complements these simulators, and delivers similar functionality such as providing an interface for constructing bandit environments from offline data. Unlike these works, our package $\mathsf{AExGym}$ is designed to model some specific, practical challenges faced by experimenters running A/B tests or field experiments who are interested in deploying adaptive algorithms. We offer support for features not usually present in existing simulators, such as 
constraints, multiple metrics, different kinds of experimental objectives (e.g., cumulative regret minimization, best-arm identification, top-5 arm identification, etc.) and non-stationarity. We also provide implementations for a suite of best-arm identification algorithms, which are not typically the focus of contextual bandit or recommendation system simulators, but are of direct relevance for experimenters. 

\section{AExGym}
\label{sec:aexgym}

To promote inductive approaches to algorithm development,
we create AExGym, an open source software package that is meant to be accessible and flexible. AExGym is designed with the purpose of allowing experimenters to easily evaluate and develop algorithms in a customized setting. For the purpose of evaluation, we first provide an environment interface that can easily incorporate existing data, customized objectives and constraints. 
As we demonstrate in Section~\ref{sec:experiments}, we instantiate our modular framework to benchmark existing algorithms on realistic environments that reflect the myriad of challenges that commonly arise in practice. 

\begin{figure}[t]  
    \centering
    \includegraphics[width=\linewidth]{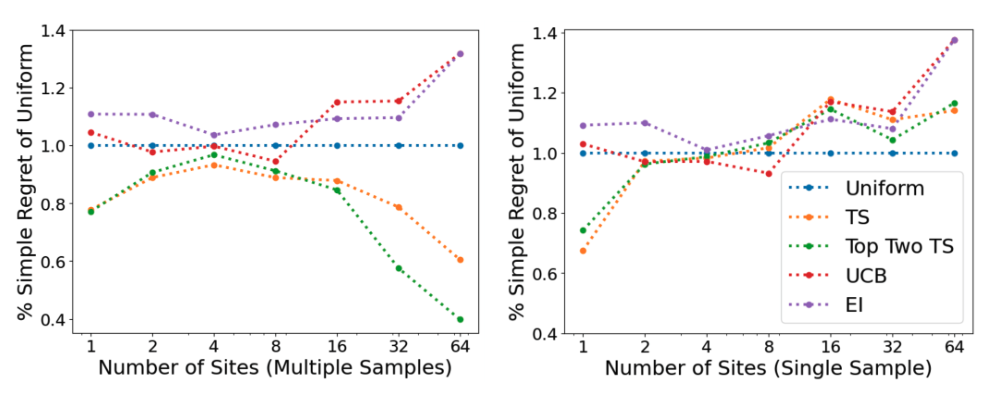}  
    \caption{Illustration of performance degradation under constraints in a site selection task constructed from the \citet{Meager2019Understanding} multi-site study. \textbf{(Left)} When each arm (site) can be sampled multiple times, Thompson Sampling based algorithms perform relatively well. \textbf{(Right)} When arms can only be sampled once (which is a common practical constraint), all algorithms perform worse than uniform.}
    \label{fig:select-sites}
\end{figure}

\subsection{Environment}

Adaptive experiments involve a sequential decision making process where at each step, an experimenter makes a decision based on existing information, and receives some form of tangible feedback from an environment. The experimenter is interested in collecting data on an outcome $R(X_{i},A_{i})$, which depends on the context of the sampling unit $X_{i} \in \R^{p}$ as well as one of $K$ assigned treatments $A_{i} \in [K]$. We denote $r(x,a)$ to be the conditional mean reward $r(x,a) := \E\bigl[R(X_{i},A_{i})\,|\, X_{i} = x, A_{i} = a\bigr]$.
Note that $\mathsf{AExGym}$ also provides support for multiple outcomes, where $R(x,a)$ is vector-valued, but focus on a single outcome for now for clarity of presentation.


We denote the adaptive experiment environment as the $\env$, which is composed of $T$ sequential epochs (horizons). Typically, we imagine $T$ to be small (< 10), in stark contrast to fully interactive bandit environments. Within each epoch $t$, a batch of $n_t$ sampling units, each with a context $X_{i}^{t}$ arrive and are assigned to treatment arms $A_{i}^{t}\in [K]$ by an adaptive algorithm. After treatment assignment, outcomes $R_{i}^{t} = R(X_{i}^{t}, A_{i}^{t})$ are observed. The adaptive algorithm, denoted as the $\mathsf{Agent}$, assigns treatments according to a probability distribution $\pi(\,\cdot\, |\,X_{i}, \mathcal H_{t})$, which may depend on the unit context $X_{i}^{t}$ and the history $\mathcal H_{t}$ of observed contexts, treatment assignments, and outcomes. More precisely, at each epoch $t \in [T]$ of the $\env$, the following events occur:
\begin{enumerate}
\item A batch of $n_{t}$ units arrives with contexts $\{X_{i}^{t}\}_{i=1}^{n_{t}}$,
drawn i.i.d. from a context distribution $\mu_{t}$, which may change over time.
\item The $\mathsf{Agent}$ selects a treatment assignment policy $\pi_t$ and assigns each user $X_i^t$ to a treatment according to $A^{t}_{i}\sim\pi_{t}(\,\cdot\, | \, X_{i}^{t}, \mathcal H_{t})$. 

\item A batch of outcomes $\{R_{i}^t\}_{i=1}^{n_{t}}$ corresponding to each unit in the batch
are observed and the history is updated $\mathcal H_{t+1} = \mathcal H_{t} \cup \{(X_{i}^{t}, A_{i}^{t}, R_{i}^{t}) \}_{i=1}^{n_{t}}$.
\end{enumerate}
After the $T$ epochs, there is a $\mathsf{PostExperiment}$ phase $T+1$ where the experimenter may make a final treatment selection for $N$ users drawn from the population distribution, $X^{T+1}_{i} \sim \mu$. In the case of best-arm identification the deployed treatment $A^{T+1}$ is identical across users, but in a personalization settings, $A^{T+1}_{i}$ could be specific to each user.
\begin{figure}[t]  
    \centering
    \includegraphics[width=\linewidth]{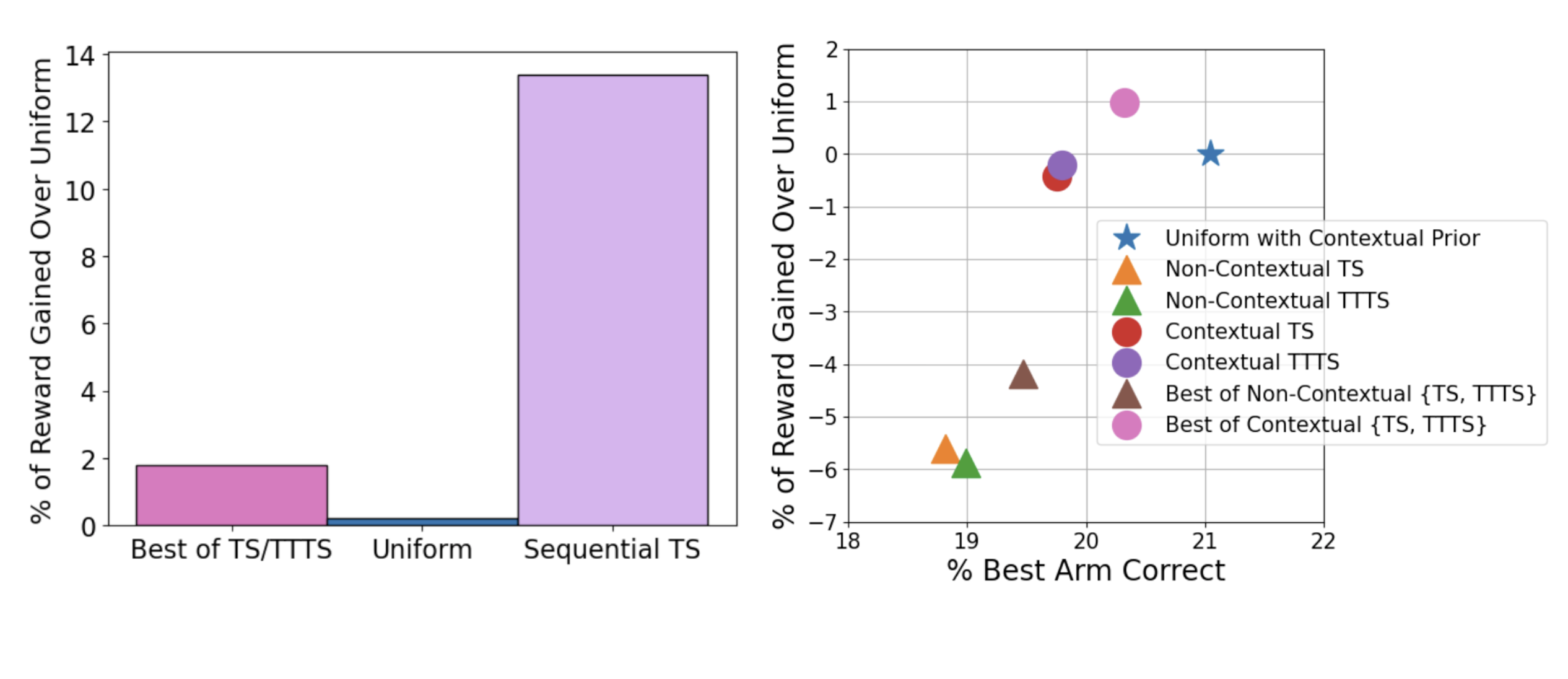}  
    \caption{Performance of various algorithms across 241 settings in the ASOS.com dataset. \textbf{(Right)} Plot of best-arm identification performance for algorithms that rely on contextual and non-contextual models. The methods tested universally perform worse than uniform due to the challenging, but naturally occurring non-stationary within the environment ($n_{t} = 100,000$).  \textbf{(Left)} Illustration of performance losses due to batching. Contextual Thompson Sampling that updates after every sample vastly outperforms Uniform despite under-performing in the batched setting ($n_{t} = 10,000$).}
    \label{fig:asos-plots}
\end{figure}
The $\mathsf{Agent}$
is evaluated based on objectives defined by the experimenter. To accommodate the diversity of objectives held by experimenters, we introduce the $\mathsf{Objective} = \{v_{t}\}_{t=1}^{T+1}$ where each $v_{t}$ is a function that assigns a value based on the history and the treatment decisions. This could include common objectives such as \emph{simple regret}, in which case $v_{T+1}$ measures the regret of the final treatment selection and $v_{t} = 0$ for $t\leq T$. It also could include more complex objectives like \emph{policy regret}, for which $v_{T+1}$ is the regret compared to the optimal policy that gives personalized treatments to each $X_{i}$ to maximize $r(X_{i},a)$:
\begin{align*}
\mathsf{SimpleRegret}\text{: } & v_{T+1}= \max_{a} \E_{X_{i} \sim \mu} \bigl[r(X_{i},a)\bigr] - \E_{X_{i} \sim \mu}\bigl[r(X_{i},A^{T+1})\bigr], \\
\mathsf{PolicyRegret}\text{: } & v_{T+1}= \E_{X_{i} \sim \mu} \bigl[ \max_{a} r(X_{i},a) \bigr] - \E_{X_{i} \sim \mu}\bigl[r(X_{i},A_{i}^{T+1})\bigr].
\end{align*}
Our flexible and modular framework can incorporate any set of outcomes/metrics; the analogous $\mathsf{Constraints}$ object can model constraints that must be satisfied by the policy. For example, a practitioner may be interested in identifying an optimal arm with respect to the primary outcome (e.g., revenue), subject to constraints that the arm does not cause a major degradation in a secondary outcome (e.g., reliability of service). 

To summarize and connect our environment back to the terminology of Markov decision processes, the \emph{state} returned by the $\env$ is the entire history $\mathcal H_t$, the \emph{action} taken by the agent is the assignment policy (which then produces treatment assignment for each sampling unit), and the \emph{rewards} are the experimenter defined evaluation objectives $v_t$ (note that from the perspective of the dynamic adaptive experimentation process, the ``rewards'' are not the individual-level outcomes).

\newtheorem{exa}[theorem]{Example}

\begin{exa}
    To elucidate upon these design choices, consider the Pennsylvania Reemployment Bonus Demonstration \cite{corson1991pennsylvania}, an experiment testing the efficacy of various reemployment bonuses on unemployed individuals. At each epoch,
    one may have access to (1) personal information $X_{i}$ of $n_{t}$ individuals that must be assigned a treatment group, (2) action features $z_{a}$ corresponding to the reemployment bonus amount, (3) costs $c(X_{i},A_{i})$ associated with giving individual $x$ a bonus $A_{i}$, and (4) distributions of future individuals $\mu_{t+1:T}$. Knowledge of future distributions could include information about future sites of experimentation (e.g., counties, schools). Finally, the outcomes $\{R(X_{i}, A_{i})\}_{i=1}^{n_{t}}$ can include the time spent on Unemployment Insurance (UI), employment, and future earnings. 
\end{exa}

\subsection{The Agent: Algorithm Design}
Typical adaptive experimentation algorithms maintain models of the environment to guide allocations decisions.
To capture this, a core feature of simulator is the $\mathsf{AgentState}_{\pi}(\mathcal H_{t})$ which condenses the history of observations into summary statistics used by sampling policy. That is, 
\[
A^{t}_{i}\sim\pi_{t}(\,\cdot\, | \, X_{i}^{t}, \mathsf{AgentState}_{\pi}(\mathcal H_{t})).
\]

For example, for a Thompson sampling (TS) policy, $\mathsf{AgentState}_{\pi}(\mathcal H_{t})$ is the posterior distribution over the treatment effects. In contextual settings, this could also involve statistical models for the rewards, such as the ordinary least squares (OLS) or logistic regression. For a few concrete examples, consider the following policies which use a linear model $r(x,a) = \phi(x,a)^\top \theta^{*}$ where $\phi(x,a)$ is a known feature mapping and $\theta^{*}$ is an unknown coefficient vector to be estimated.


\begin{figure}[h]
\begin{python}[
caption=Linear TS in a personalization environment with simple regret as an objective.,
label=fig:code-example,
]
agent = LinearTS()
objective = ContextualSimpleRegret()

# Reset environment and policy
state_contexts, action_contexts, cur_step = env.reset()
beta, sigma = agent.model.reset()

# Environment interactions
while env.n_steps - cur_step > 0:
    
    # Compute action probabilities from the agent
    action_probabilities = agent(
        beta=beta, 
        sigma=sigma, 
        state_contexts=state_contexts,
        action_contexts=action_contexts, 
        objective=objective
    )
    
    # Sample actions
    actions = torch.distributions.Categorical(action_probabilities).sample()
    
    # Take a step in the environment
    state_contexts, action_contexts, rewards, cur_step = env.step(actions)
    beta, sigma = agent.model.update_posterior(
        beta=beta,
        sigma=sigma,
        state_contexts=state_contexts,
        action_contexts=action_contexts,
        rewards=rewards
    )

# Calculate objective in post-experiment phase
post_contexts, action_contexts, post_rewards = env.get_post_experiment_eval()
exploit_actions = agent.exploit(
    beta=beta, 
    post_contexts=post_contexts,
    action_contexts=action_contexts
)
regret = objective(actions=exploit_actions, rewards=post_rewards)
\end{python}
\vspace{-2.5ex}
\end{figure}

\begin{exa}[Linear TS]
    Let $\mathsf{AgentState}_{\pi}(\mathcal H_{t}) = (\theta_{t}, \Sigma_{t})$, which parameterizes a Gaussian posterior distribution of $\theta^{*}$ given data in $\mathcal H_{t}$. The assignment policy samples from 
    \[
    \pi(\,\cdot\,|\,X_{i}, \mathcal H_{t}) = \P_{\theta \sim N(\theta_{t},\Sigma_{t})}\Bigl(\argmax_{a'} \{\phi(x,a')^\top \theta \} = a\Bigr), 
    \] which is the posterior probability that action $a$ is optimal. See Figure \ref{fig:code-example} for a code example of Linear TS in an environment.
\end{exa}

\begin{exa}[Linear UCB]
    Let $\mathsf{AgentState}_{\pi}(\mathcal H_{t}) = (\hat{\theta}_{t}, V_{t})$, where $\hat{\theta}_{t}$ is the OLS estimator using data in $\mathcal H_{t}$ and $V_{t}$ is the (regularized) design matrix of $\hat{\theta}_{t}$. Linear UCB chooses an action as 
    \[
    \pi(\, \cdot \, |\,X_{i}, \mathcal H_{t}) = \argmax_{a} \Bigl \{\phi(x,a)^\top \hat{\theta}_{t} + \alpha_{t} \sqrt{\phi(x,a)^\top V_{t}^{-1} \phi(x,a)} \Bigr \}\] 
    for a choice of $\alpha_t > 0$. 
\end{exa}

$\mathsf{AgentState}$ is designed to accommodate general statistical models (e.g., linear models or even a neural network), and enables comparison between the same adaptive algorithm under different statistical procedures. 
\section{Datasets}
\label{datasets}
To benchmark adaptive experimentation algorithms with respect to challenges most similar to those faced by practitioners, we curate a set of observational and experimental data sources reflecting realistic experimental settings. In addition to the specific challenges that each setting entails, we study the effects of batched feedback, various constraints, and multiple outcomes and objectives across all of these settings. 
\begin{enumerate}[leftmargin=*]
    \item \textbf{Meager}~\cite{Meager2019Understanding}: An analysis of $7$ Randomized Control Trials carried out in various countries for the purpose of determining the efficacy of microcredit financing. Due to resource and time constraints, experimenters may only be able to choose a few sites of experimentation at various locations. To simulate this process, we generate realistic location clusters based on three covariates within the data: site, cluster-id, and district. Clustering by these features yields $7$, $77$, and $95$ clusters respectively. We focus on $4$ different outcomes, profits, expenditures, revenues, and consumption, as a result of the treatment. We use this data to benchmark choosing the sites with the highest Average Treatment Effect, external validity, and constraints. 
    \item \textbf{Pennsylvania Reemployment Bonus Demonstration} \cite{corson1991pennsylvania}: A study testing the efficacy of $6$ different reemployment bonus programs for the purpose reducing reliance on Unemployment Insurance. We focus on the outcome of time continued on Unemployment Insurance, where lower is better. There are two metrics with regards to this outcome. Since each individual has an outcome associated with one specific treatment, we choose to fill in counterfactuals by fitting and predicting with a linear model. Using more sophisticated Off Policy Evaluation methods is of future interest. Using this data, we benchmark personalization and multiple objectives.  
    \item \textbf{National Health Interview Survey (NHIS)} \cite{NHIS2010-2019}: Similar to \cite{JeongNa22}, we use the NHIS to estimate the effect of Medicaid enrollment on doctors' office utilization. We use a proxy for this outcome by observing whether the subject visited a doctors' office two weeks before the survey date. We generate clusters within the data by utilizing the covariates indicating the country of birth, yielding $12$ different clusters. We benchmark the site selection task on this data. 
    \item \textbf{ASOS Digital Experiments} \cite{liu2021datasets}: There are $78$ online A/B tests within this dataset. Each A/B test is associated with multiple metrics. We treat each metric as a separate experiment, yielding $241$ different settings. The data consists of means and variances at various time steps, which are highly non-stationary. We use this dataset to benchmark algorithms in a non-stationary environment with realistic treatment effect sizes. 
    \item \textbf{Field Experiments}: We collect means and variances from $16$ educational multi-site experiments \cite{weiss2017effects, wahlstrom2014examining}, $3$ welfare-to-work studies \cite{Riccio1992GAIN, Hamilton2001National, Kemple1995Florida}, and an experiment involving law enforcement \cite{ Ariel2016Wearing}. These means and variances are reported within the papers and reports mentioned above. A few of the reports do not have variances and instead have markers of statistical significance, so we use these markers to estimate the variance of the statistic. Since these means are highly non-stationary, we benchmark algorithms on non-stationarity and realistic treatment effect sizes. 
\end{enumerate}

\section{Experiments}
Using these datasets, we explore several case studies, which reveals that the practical performance of bandit algorithms can be greatly impacted by operational constraints that are seldom studied in bandit literature. Within these constraints, often simpler sampling rules such as uniform sampling may have more desire-able properties. 
More specifically, we explore the performance of algorithms under challenges including personalization, non-stationarity, external validity, multiple objectives, and batched feedback. 
We provide a number of implementations of standard bandit and best-arm identification policies. These appear in several of the benchmark tasks we consider. We detail these policies below. 

\begin{figure}[t]
    \centering 
    \includegraphics[width=0.6\textwidth]{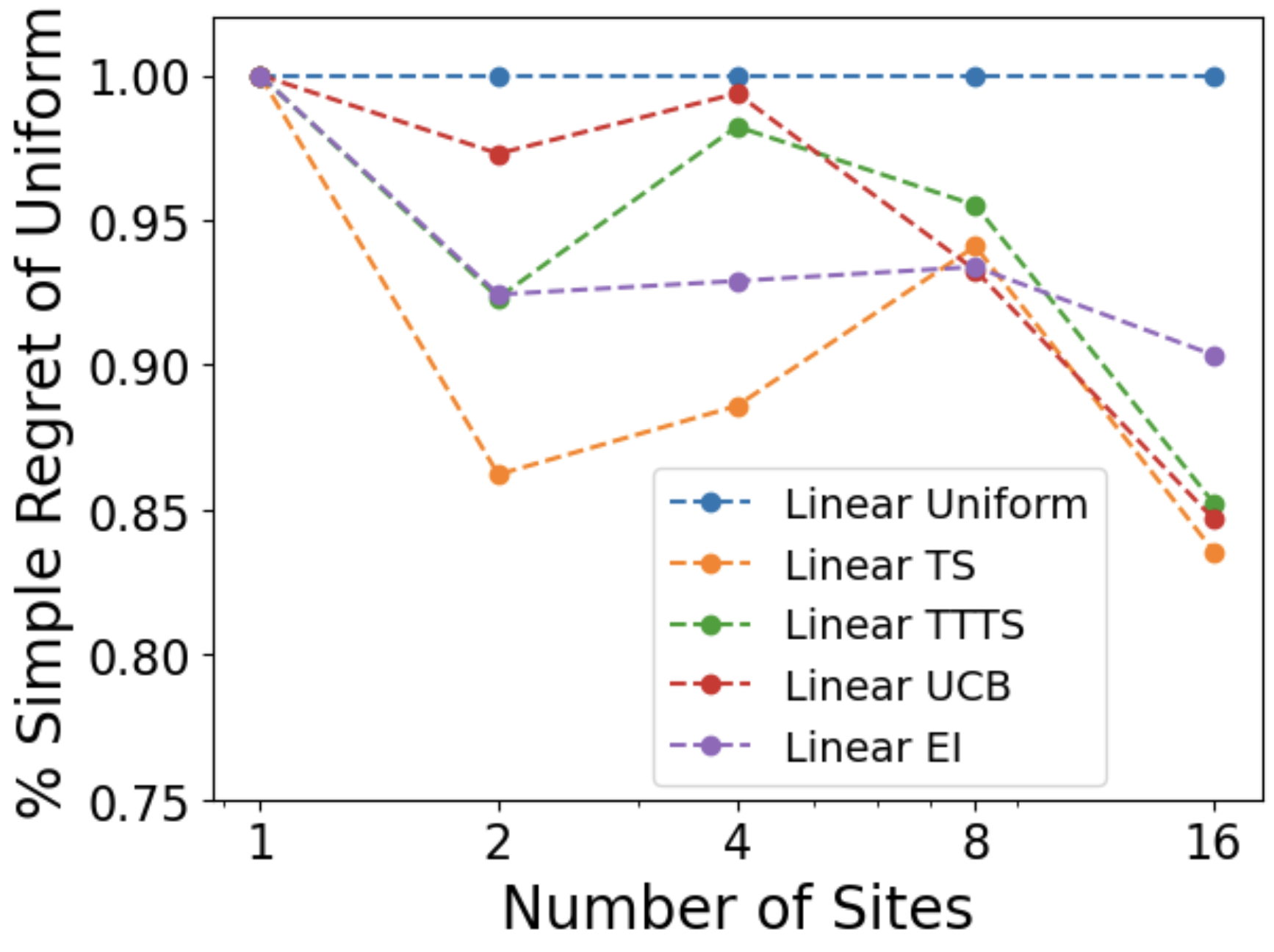}
    \caption{Outcomes for a site selection task for the NHIS data. All algorithms generally do better than the Uniform allocation. }
    \label{fig:nhis-site}
\end{figure}

\begin{itemize}
    \item \textbf{Uniform}: Uniformly allocates samples across all arms. 
    \item \textbf{Linear Thompson Sampling (TS)} \cite{agrawal13thompson}: Thompson Sampling under a linear-contextual model.
    \item \textbf{Linear Top-Two Thompson Sampling (TTTS)}: Top-Two Thompson Sampling under a linear-contextual model. For the simple regret objective, TTTS is exactly Deconfounded TS described in~\cite{QinRusso2023}. 
    \item \textbf{Linear Upper Confidence Bound (UCB)}~\cite{Abbasi-YadkoriPaSz11}: For each unit, selects the treatment with the highest upper confidence bound under a linear model.
    \item \textbf{Linear Expected Improvement (EI)}~\cite{jones1998efficient}: For each unit, selects the treatment that maximizes the expected improvement index under a linear model.
\end{itemize}
\label{experiments}

\subsection{Selecting Sites with Large Effects}
When evaluating treatments in large-scale field experiments, researchers are often interested in selecting sites where interventions will have large effects. Consider a local government that wishes to allocate limited funds to various school districts. Ideally, they would like to allocate funds to poor, necessitous school districts that may benefit a large amount. However, due to the costs associated with these treatments, researchers can only run experiments for a limited number of sites. By sequentially rolling out treatments in different sites, adaptive algorithms can use observed treatment effects to identify promising sites, using similarity in site-level features to guide exploration. To benchmark different algorithms for this task, we use data from \citet{Meager2019Understanding}, which details outcomes from microfinance studies in 7 countries. Each country has several districts, and we filter down to 77 country-district pairs that experienced both treatment and control, which we treat as the sites.

We model this problem in our framework as an adaptive experiment where `treatments' are the site selections, so we have that $A_{i} \in [K]$ for $K$ sites. We consider a setting where there are $T$ epochs with $T < K$ and a single sampling decision in each epoch (so the batch size $n_{t} = 1)$.
The outcome $R(X_{i}, A_{i})$ of interest are the average treatment effect (ATE) for the site, which is modeled as a linear function of the site characteristics $X_{i}$, with a coefficient $\theta$ common across sites:
\[
r(X_{i}, A_{i}) = \theta^{\top} X_{i}
\]
Whenever a site is sampled, $R(X_{i}, A_{i})$ is generated by a bootstrapped sample of the average treatment effect using household-level outcomes, reflecting the level of noise in the ATEs. At the end of the experiment, the experimenter makes a final site selection $A_{T}$ that represents their belief of the site with the highest ATE. The $\mathsf{Objective}$ is to minimize $\mathsf{SimpleRegret}$ compared to the site with true highest ATE. 

We benchmark Linear Thompson Sampling~\cite{agrawal13thompson}, Linear Top-Two Thompson Sampling~\cite{Russo20}, Linear Upper Confidence Bound (UCB)~\cite{Abbasi-YadkoriPaSz11}, Expected Improvement~\cite{QinKlRu17}, all drawn from the bandit and Bayesian Optimization literatures.

\begin{table}[t]
\centering
(a) Regret for 64 Rounds
\vspace{1em}

\begin{tabular}{lcccc}
\toprule
 & \textbf{profit} & \textbf{revenues} & \textbf{expenditures} & \textbf{consumption} \\
\midrule
Uniform & 0.537 & 0.761 & 0.483 & 0.751 \\
TS & 0.638 & 0.777 & 0.487 & 0.777 \\
TTTS & 0.572 & 0.842 & 0.582 & 0.804 \\
UCB & 0.621 & 0.834 & 0.562 & 0.802 \\
EI & 0.640 & 0.815 & 0.602 & 0.810 \\
\bottomrule
\end{tabular}

\vspace{1em}
(b) Regret for 16 Rounds
\vspace{1em}

\begin{tabular}{lcccc}
\toprule
 & \textbf{profit} & \textbf{revenues} & \textbf{expenditures} & \textbf{consumption} \\
\midrule
Uniform & 0.512 & 0.778 & 0.566 & 0.832 \\
TS & 0.593 & 0.785 & 0.555 & 0.754 \\
TTTS & 0.593 & 0.783 & 0.554 & 0.791 \\
UCB & 0.604 & 0.804 & 0.581 & 0.802 \\
EI & 0.592 & 0.835 & 0.576 & 0.831 \\
\bottomrule
\end{tabular}

\vspace{1em}

\caption{Comparison of regret for the site selection task for 95 sites using the Meager~\cite{Meager2019Understanding} data. (a) Regret across several metrics (profit, revenues, expenditures, and consumption) under 64 sequential rounds. (b) Regret across metrics under 16 sequential rounds.}
\label{tab:meager-site-selection}
\end{table}

The flexibility of our framework allows us to explore natural constraints around this problem. Figure~\ref{fig:select-sites} compares the performance these algorithms relative to the $\mathsf{SimpleRegret}$ incurred by the uniform sampling policy. The left panel allows for sites to be sampled multiple times, while the right panel details performance under the natural constraint that the site may only be sampled once. This is almost always the case in practice in field experiments.

There are a couple notable observations. First, unlike UCB or EI, the Thompson Sampling algorithms are able to explore under limited sampling and can substantially reduce the simple regret by revisiting previously sampled sites. However, under the constraint, this advantage vanishes and all the policies surprisingly under-perform uniform sampling as the sampling budget $T$ grows large. Contrary to the idealized settings studied in theory, operational constraints on sampling can impact  practical performance, in unexpected ways. This further demonstrates the value of simple algorithm like uniform sampling, in navigating these constraints.

Table~\ref{tab:meager-site-selection} show results for the site selection task with $95$ sites. The top panel details the results when the experimenter can select up to $64$ sites and the bottom panel shows the results when only $16$ sites can be sampled. 
The experimenter sees a noisy observation of the average treatment effect (ATE) in each sampled site across $4$ outcomes we measure in the data: profit, revenues, expenditures, and consumption. After sampling the sites, the experimenter selects a site that maximizes the metric, and is evaluated based on the regret compared to the site with the highest ATE. Surprisingly, we observe that uniformly sampling sites results in lower regret compared to more sophisticated sampling methods. This holds for both sampling budget sizes.

We also consider the site selection task under the NHIS survey data~\cite{NHIS2010-2019}. The treatment in question is Medicaid enrollment and the main outcome is whether the subject visited a doctors' office 2 weeks before the survey date. We consider 12 clusters of study participants, based on covariates such as the country of birth.  The task is to identify the cluster that has the highest treatment effect. Once the experimenter samples a cluster, they see a noisy observation of the treatment effect. Figure \ref{fig:nhis-site} shows the results. As opposed to the previous site selection task, we observe that all the adaptive sampling methods reduce regret compared to the uniform sampling policy. Thompson Sampling, Top-Two Thompson Sampling, and UCB all perform similarly.

\subsection{Nonstationarity}

Another practical concern of experimenters is non-stationarity. To explore this issue, we conduct a rigorous empirical study across 241 unique settings within the ASOS Digital Experiments Dataset. This dataset contains results from 78 real-world experiments 
 carried out by a business unit within ASOS.com, a global fashion retailer with over 
 26 million active customers (as of 2024). Each experiment includes up to four different metrics. Although the dataset does not contain individual-level rewards, it consists of  the sample means $\theta_{t,a}$ and variances $s_{t,a}^{2}$ of a control arm and a treatment arm recorded at either 12-hour or daily intervals. Our simulation framework can also work flexibly with using moment or distributional-level data to generate arm rewards.

 Reflecting existing experimental practice, we consider a batched experiment over $T = 10$ epochs, which are the 12-hour or daily intervals. Each batch consists of $n_{t} = 10,000$ units. Using our framework, we generate additional treatment arms which are similar to the treatment outcome, so that there are $K = 10$ arms total. The user feature is simply the epoch $t$ they arrive to the experiment. Rewards are $R(t,a) \sim N(\theta_{t,a}, s_{t,a}^{2})$, so that the mean rewards are
 \[
r(t,a) = \theta_{t,a}.
 \]
 We observe in the data that $\theta_{t,a}$ fluctuates greatly across days, and the treatment with the highest outcome actually changes day-by-day. As a result, it is natural for the experimenter to be interested in identifying the treatment with the highest outcome when averaged over the duration of the experiment:
 \begin{equation}
 \bar{r}(a) = \frac{1}{T} \sum_{t=1}^{T} \theta_{t,a}
 \end{equation}
 At the end of the experiment, the experimenter makes an arm selection.
 The $\mathsf{Objective}$ of the experimenter is to minimize the simple regret with respect to $\max_{a} \bar{r}(a)$.

 \begin{figure}[t]
    \centering
    \includegraphics[width=0.6\textwidth]{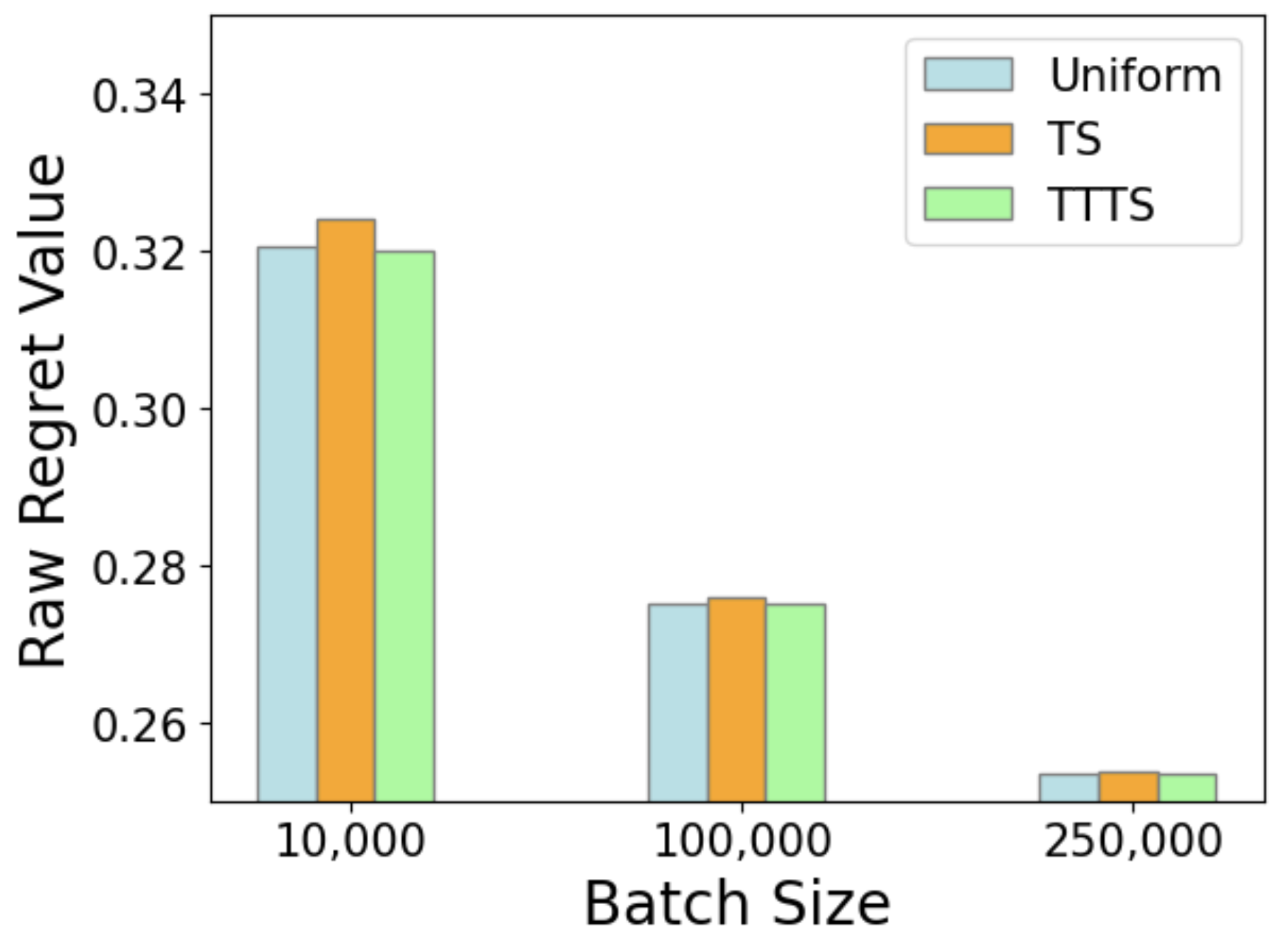}
    \caption{Raw regret values for the ASOS data with batch sizes of $10,000$, $100,000$, $250,000$. Regret values decrease as batch sizes get larger. }
    \label{fig:nonstationary-regret}
\end{figure}

 We benchmark standard batched variants of multi-armed bandit Thompson Sampling (TS) and Top-Two Thompson Sampling algorithms (TTTS), as well as batched contextual TS (Context-TS) and TTTS (Context-TTTS) algorithms~\cite{QinRusso2023} that can use temporal features to be robust to non-stationarity. As has been pointed out in~\cite{QinRusso2023}, non-stationarity can be highly detrimental to the performance of non-contextual bandit algorithms, which severely underperform uniform sampling as can be seen in the right panel Figure~\ref{fig:asos-plots}. Yet even the contextual policies experience a larger regret than uniform, albeit slightly more. This analysis reveals another surprising observation. If we removed the batch constraint, then the contextual Thompson Sampling policy substantially outperforms uniform sampling as seen in the left panel of Figure~\ref{fig:asos-plots}. This illustrates that batching does indeed impose a significant constraint on the capabilities of adaptive algorithms. While raw regret values decrease as batch size increases as shown in Figure~\ref{fig:nonstationary-regret}, TS policies are unable to outperform Uniform by a large amount unless they are allowed to update frequently.  

 We further run Contextual TTTS and non-Contextual TS on the field experiment data with the number of batches equating the number of sites within the data, $10$ arms, and with batch sizes specified by the paper. We find that the Contextual TTTS method does $0.5\%$ better the Uniform, while the non-contextual method does $1.8\%$ worse than Uniform. This reinforces the idea that non-stationarity can significantly impact the performance of adaptive methods.

\subsection{Multiple Objectives}
\label{subsection: multiple-objectives}
\begin{figure}[h]
    \centering
    \includegraphics[width=0.6\textwidth]{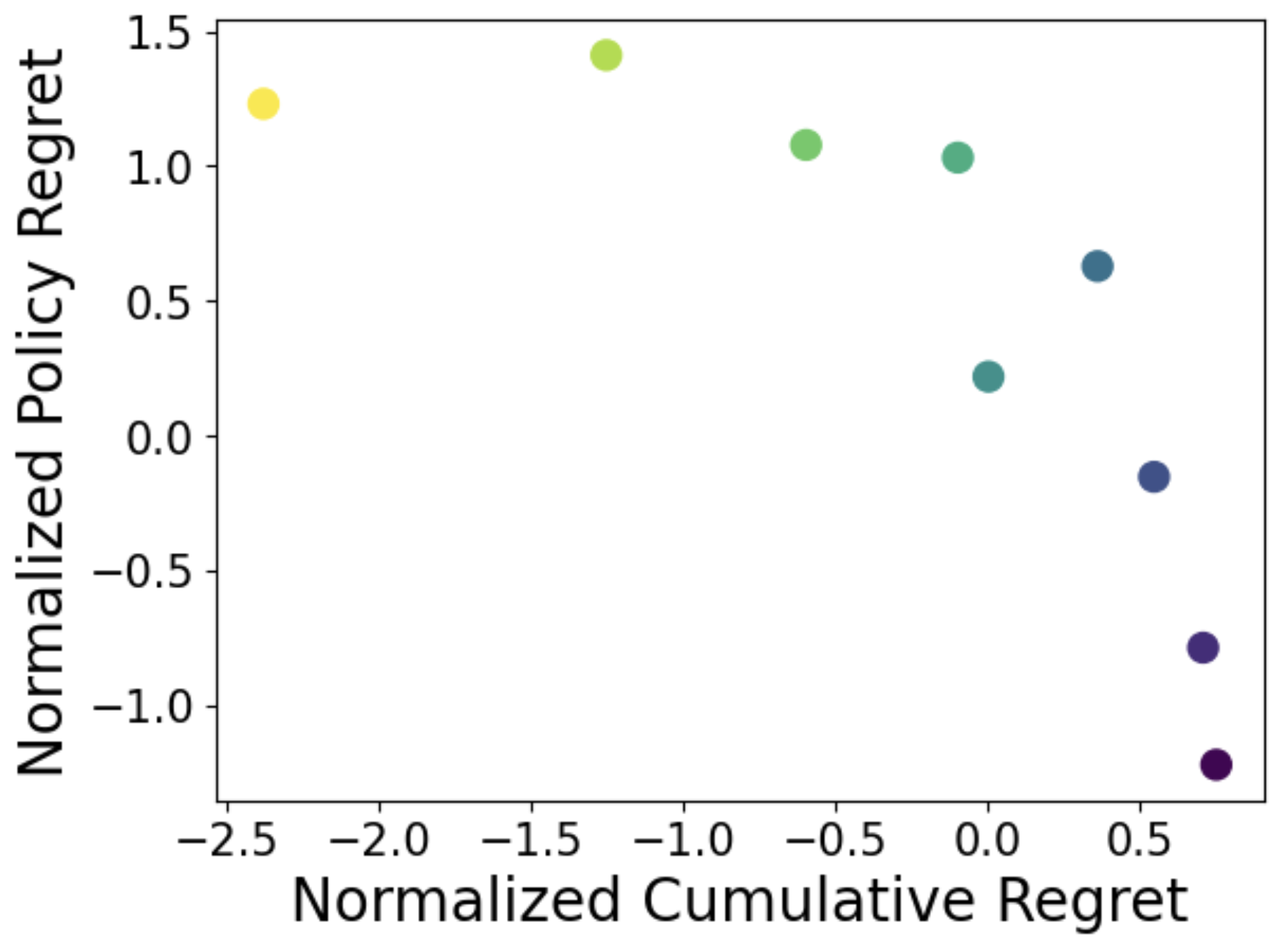}
    \caption{Pareto frontier illustration of the tradeoff between simple and cumulative regret within the Pennsylvania Reemployment Bonus Demonstration data. We vary the parameter of Top Two Thompson Sampling from $0.2$ to $1$ by increments of $0.1$. Dark blue corresponds to $0.2$, while lighter colors indicate higher valued parameters with yellow being $1$. Notably, the lightest point with parameter $1$ is exactly the original Thompson Sampling, and incurs the smallest cumulative regret. Parameters closer to $0.2$ incur less policy regret.}
    \label{fig:simple-cumul-regret}
\end{figure}

In practice, experiments often involve multiple competing outcomes and objectives that experimenters wish to optimize. One example is the tradeoff between the regret incurred within the experiment, and the regret incurred after the experiment. During the experiment, the experimenter wishes to minimize cumulative regret and maximize the welfare of the participants. On the other hand, the experimenter wishes to explore various treatment arms in order to determine the best treatment arms after the experiment. These objectives are often at odds with one another, as shown in Figure \ref{fig:simple-cumul-regret}. This figure depicts the varying performance of Top Two Thompson Sampling on the Pennsylvania Reemployment Bonus Demonstration data with $5$ batches of $10,000$ samples. Top Two Thompson Sampling involves a parameter that ranges from $0$ to $1$, controlling its level of greediness. A setting of $1$ corresponding exactly to the original Thompson Sampling algorithm which is more greedy. The figures shows that parameters closer to $1$ (lighter) incur less cumulative regret but more policy regret, while parameters closer to $0$ (darker) incur less policy regret but more cumulative regret. 

One can also use our environments to benchmark various algorithms' performance with respect to multiple outcomes within the data. The Meager data~\cite{Meager2019Understanding} involves $4$ different outcomes, the Pennsylvania Reemployment Bonus Demonstration data involves $2$, and the ASOS data has multiple metrics corresponding to each experiment. Although we do not benchmark algorithms in this aspect, our simulation setup is able to accomodate this setting easily.

\begin{table}[h!]
\centering
\begin{tabular}{lcccccccc}
\toprule
& \multicolumn{2}{c}{\textbf{profit}} & \multicolumn{2}{c}{\textbf{revenues}} & \multicolumn{2}{c}{\textbf{expenditures}} & \multicolumn{2}{c}{\textbf{consumption}} \\
\cmidrule(lr){2-3} \cmidrule(lr){4-5} \cmidrule(lr){6-7} \cmidrule(lr){8-9}
 & \textbf{single} & \textbf{multi} & \textbf{single} & \textbf{multi} & \textbf{single} & \textbf{multi} & \textbf{single} & \textbf{multi} \\
\midrule
Uniform & 0.580 & 0.617 & 0.780 & 0.761 & 0.484 & 0.547 & 0.820 & 0.792 \\
TS & 0.606 & 0.649 & 0.787 & 0.796 & 0.506 & 0.546 & 0.803 & 0.724 \\
TTTS & 0.575 & 0.616 & 0.844 & 0.819 & 0.554 & 0.524 & 0.755 & 0.693 \\
UCB & 0.574 & 0.675 & 0.790 & 0.771 & 0.582 & 0.522 & 0.741 & 0.720 \\
EI & 0.645 & 0.654 & 0.823 & 0.824 & 0.551 & 0.539 & 0.747 & 0.728 \\
\bottomrule
\end{tabular}
\vspace{1em}
\caption{Regret across metrics for the site selection task for the Meager~\cite{Meager2019Understanding} dataset when algorithms are constrained to select each site once (single) versus multiple times (multi). Performance is mixed across algorithms and there is no clear trend. }
\label{tab:site-constraint}
\end{table}

\subsection{Constraints}
\label{section:constraints}

Experimenters often deal with practical constraints including geographical and budget constraints. As shown in Section \ref{experiments} and Figure \ref{fig:select-sites}, when the site selection task is restricted such that algorithms may only choose each site once, algorithms that perform better originally may perform worse under these new constraints. Building on this work, Table \ref{tab:site-constraint} measures performance across $4$ outcomes in the Meager~\cite{Meager2019Understanding} under these constraints. The data is clustered by district, and performance is mixed in this setting. There is no clear trend as to when algorithms will perform better when able to choose a site multiple times versus a single time.  

\begin{figure}[t]
    \centering
    \includegraphics[width=0.6\textwidth]{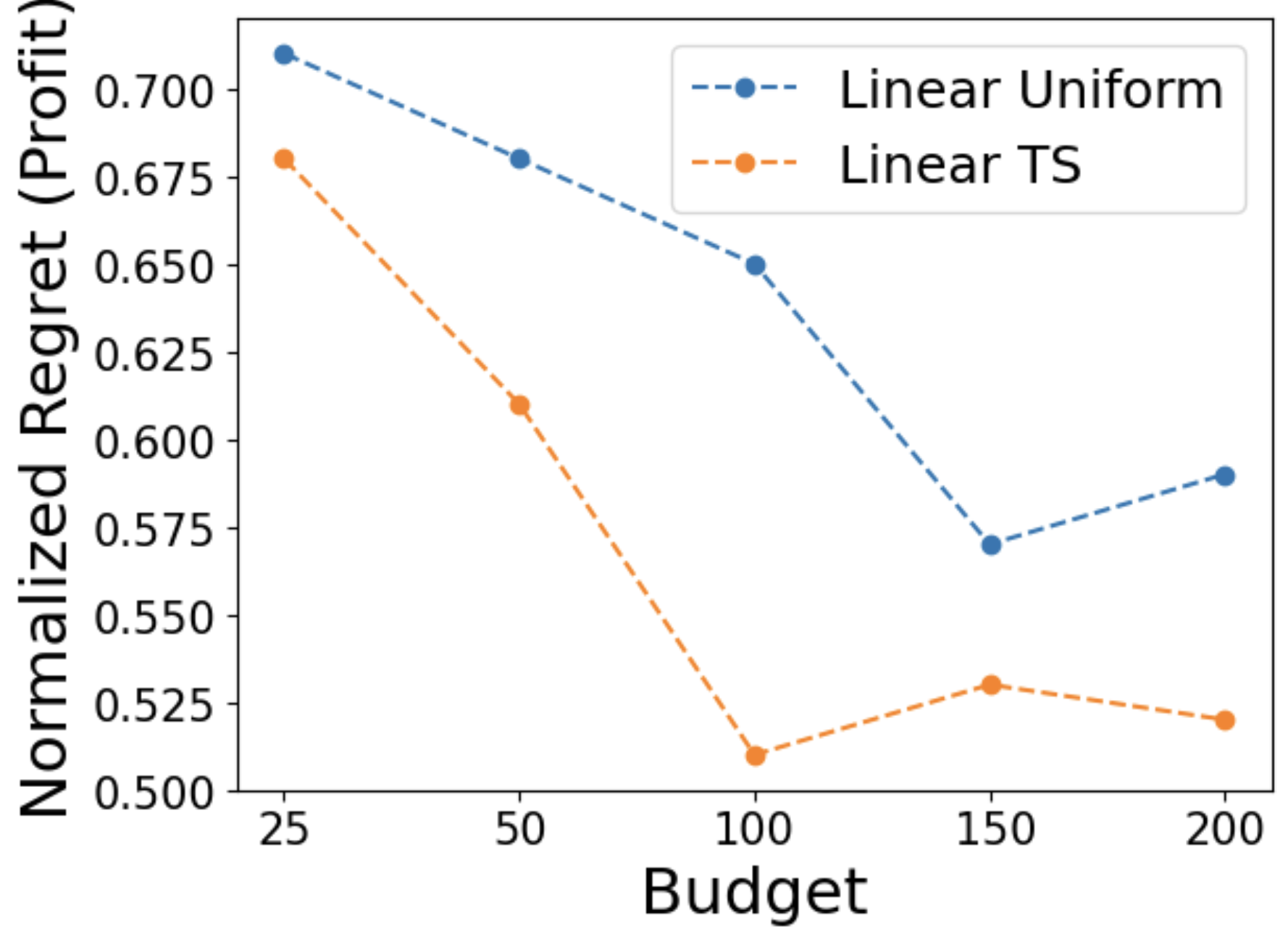}
    \caption{Regret values for Meager data when each site is associated with a cost $c_{a}$. Constrained TS does well by choosing cheaper sites that are worth exploring, while Uniform is unable to explore as it randomly chooses sites, which may be expensive. }
    \label{fig:budget-constraint}
\end{figure}

In addition to geographical constraints, experimenters often have budget constraints that they must consider. To simulate this setting, we use the site selection task in the Meager~\cite{Meager2019Understanding} dataset and assign a random cost $c_{a}$ to each site sampled from a normal distributed with mean $20$ and variance $10$.  We compare Uniform selection with a version of Thompson Sampling specialized for budget constraints \cite{xia2015thompson}. Given a belief about the mean of each site, this constrained Thompson Sampling samples a site proportional to $\theta_{a}^\top X / c_{a}$, which is the belief about the mean reward of each site divided by its cost. Figure \ref{fig:budget-constraint} shows the superior performance of the constrained Thompson Sampling under such budget constraints. While Uniform naively samples various sites randomly regardless of the cost associated, constrained TS is able to sample many more sites due to awareness of the cost. 

\subsection{External Validity}
\label{section: external-validity}

While experiments are typically focused on measuring outcomes at the experiment site, experimenters are typically also interested in \emph{external validity}, the ability of findings to generalize to regions not covered by the experiment. For example, experiment results from an experiment carried out in the wealthier United States may not generalize to a less developed country. 

\begin{figure}[t]
    \centering 
    \includegraphics[width=0.6\textwidth]{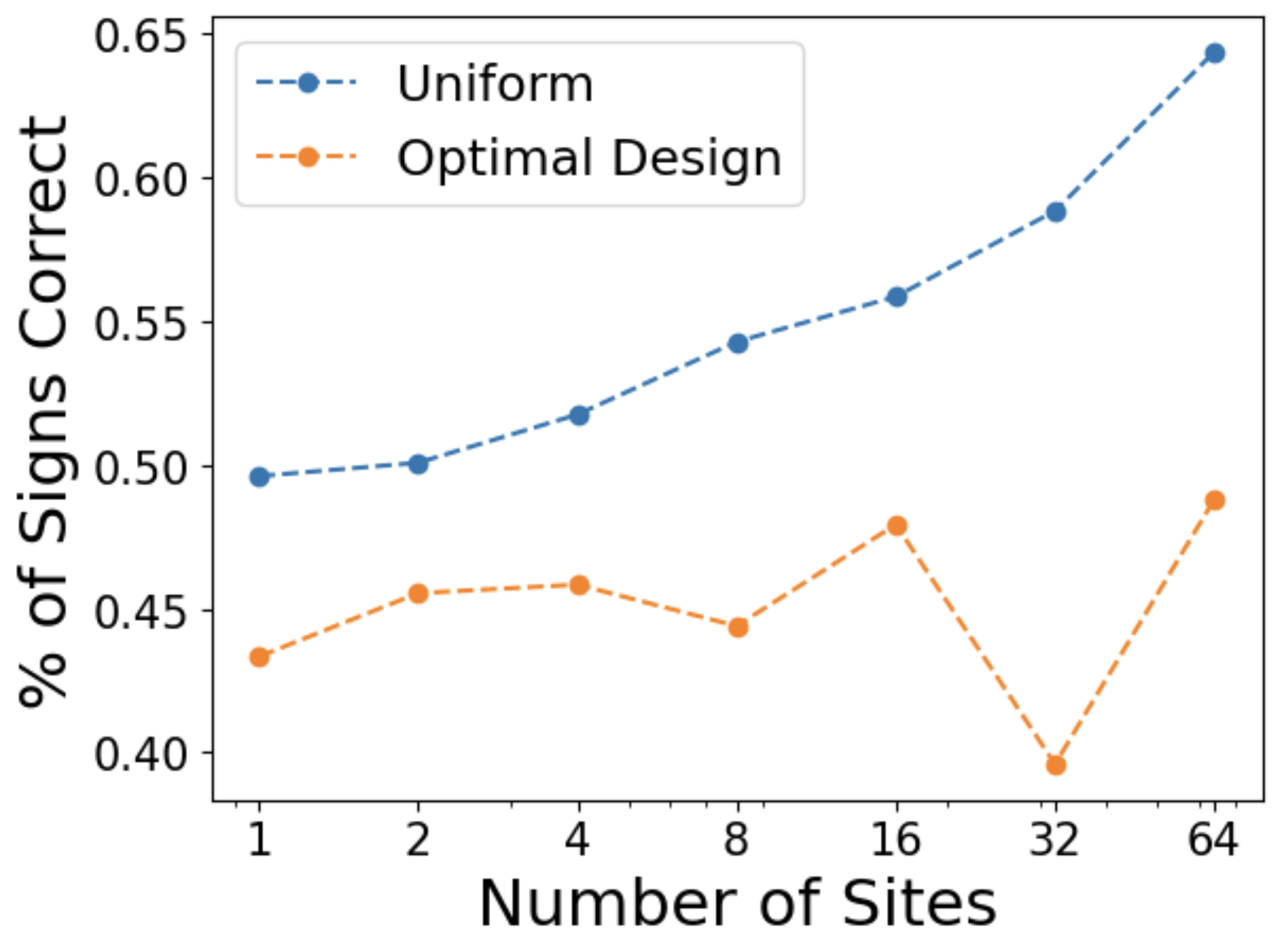}
    \caption{Percent of ATE signs predicted correctly under Uniform sampling versus an Optimal Design method after observing only a few sites.
    Surprisingly, the Uniform method outperforms the Optimal Design.}
    \label{fig:oed-external}
\end{figure}

To simulate the process of selecting sites for the purpose of external validity, we use the Meager~\cite{Meager2019Understanding} site selection task. Instead of selecting the sites with the highest Average Treatment Effect, the task is to select sites such that the experimenter can guess the signs of the ATE at sites not measured, a task known as \emph{Sign Generalization} \cite{EgamiHa22}. 

We benchmark Optimal Experimental Design (OED) methods, specifically the D-Optimal design against uniformly selecting sites for the task of Sign Generalization. D-Optimal methods select covariates with the objective of minimizing the determinant of the inverse of the the $X^\top X$ matrix. We let the D-Optimal method see the features that make up each site, and sequentially choose sites that optimize this objective.

Surprisingly, in Figure \ref{fig:oed-external}, the OED method performs worse than naively selecting sites at random. While the Uniform method steadily increases in performance as the number of sites it observes increases, the OED method stagnates. This shows that experimenters need to be careful and consider how the algorithmic method interacts with the environment. 

\subsection{Personalization}
\label{section: personalization}

\begin{figure}[t]
    \centering
    \includegraphics[width=0.99\textwidth]{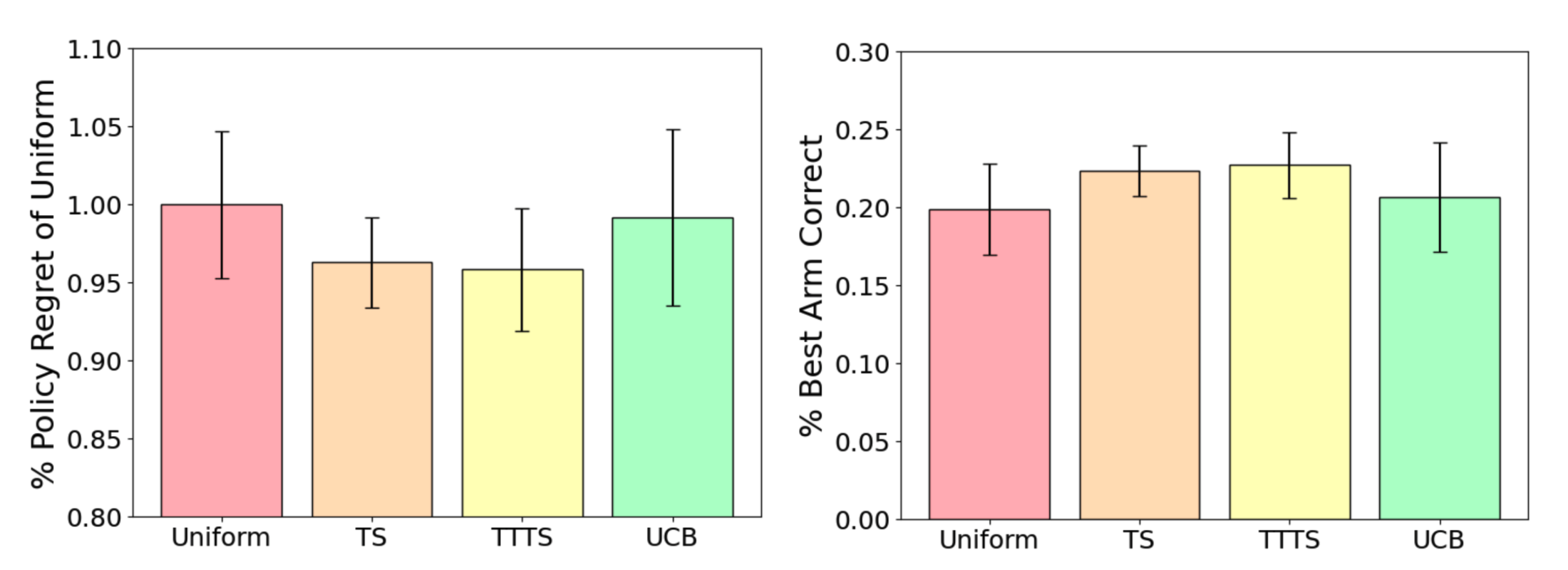}
    \includegraphics[width=0.99\textwidth]{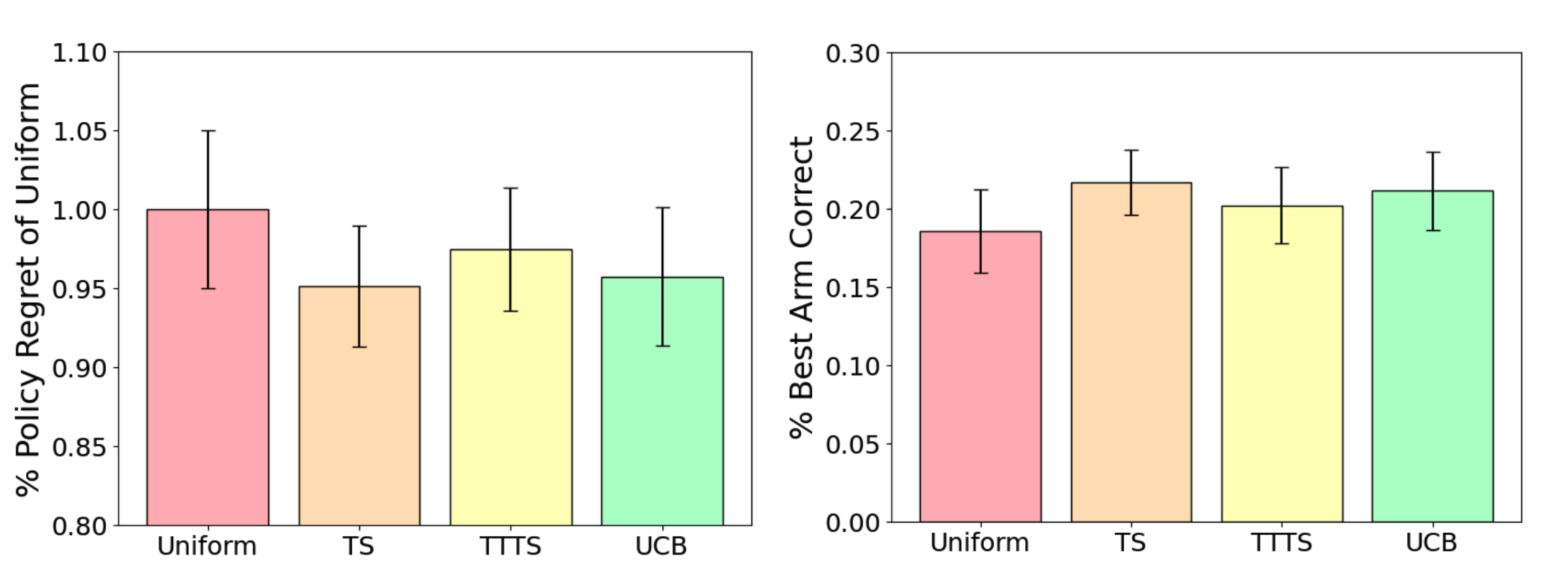}
    \caption{Comparison of policy regret for a personalization task on the Pennsylvania Reemployment Bonus Demonstration dataset. (Top) This setting involves $5$ batches of $10,000$ individuals. Thompson Sampling policies perform better both in terms of regret and best arm identification, while $UCB$ slightly outperforms Uniform. (Bottom) Results for $10$ batches of $5,000$ individuals. Notably, UCB and TS perform better than TTTS by a small margin, unlike in the setting with $5$ batches of $10,000$ individuals.}
    \label{fig:penn_10000_5}
\end{figure}

Experiments may be interested in deploying policies that are tailored to each individual. This task is motivated by the idea that individuals may react differently to various policies. To benchmark personalization experimentation methods, we create an environment from the Pennsylvania Reemployment Bonus Demonstration dataset. The data includes individual covariates $x$ and $6$ treatment arms. We simulate two settings: one with $5$ batches of $10,000$ individuals  and one with $10$ batches of $5,000$ individuals. After the experiment, the experimenter uses the collected data to form a final treatment allocation policy. The experimentation algorithm is evaluated based on $\mathsf{Policy Regret}$, the regret incurred by the final policy: $v_{T+1}= \E_{X_{i} \sim \mu} \bigl[ \max_{a} r(X_{i},a) \bigr] - \E_{X_{i} \sim \mu}\bigl[r(X_{i},A_{i}^{T+1})\bigr]$. 

We focus on linear policies that model rewards as 
\begin{equation}
    r(X_{i},A_{i}) = \theta_{A_{i}}^\top X_{i}
\end{equation}
where $A_{i}$ is one out of the $6$ treatment arms and $\theta_{A_{i}}$ is the linear coefficient associated with that treatment arm.  Linear policies make the final allocation decision at time $T+1$ by choosing $a_{i} = \argmax_{a} \theta_{T, a}^\top X_{i}$. 


We benchmark Linear Thompson Sampling, Linear Top Two Thompson Sampling, Linear UCB, and the Uniform policies. Figure~\ref{fig:penn_10000_5} shows bar charts with confidence intervals for both settings. The top panel displays the results for $5$ batches with $10,000$ individuals per batch, while the bottom panel shows results with $10$ batches with $5,000$ individuals per batch. This is to compare frequent but smaller batches with less frequent and larger batches. These figures showcase both percent regret of the Uniform policy, and also the percentage of the times the policy identifies the best arm for each individual. We observe that when there are $5$ batches, Top-Two Thompson Sampling does slightly better than other methods. However under $10$ batches, Thompson Sampling and UCB, which are designed to minimize cumulative regret, are slightly more performant for best-arm identification objectives.

\newpage 
{
\bibliographystyle{abbrvnat}
\bibliography{adaptive-bib, refs}

\begin{thebibliography}{77}
\providecommand{\natexlab}[1]{#1}
\providecommand{\url}[1]{\texttt{#1}}
\expandafter\ifx\csname urlstyle\endcsname\relax
  \providecommand{\doi}[1]{doi: #1}\else
  \providecommand{\doi}{doi: \begingroup \urlstyle{rm}\Url}\fi

\bibitem[Abbasi-Yadkori et~al.(2011)Abbasi-Yadkori, P{\'a}l, and
  Szepesv{\'a}ri]{Abbasi-YadkoriPaSz11}
Y.~Abbasi-Yadkori, D.~P{\'a}l, and C.~Szepesv{\'a}ri.
\newblock Improved algorithms for linear stochastic bandits.
\newblock In \emph{Advances in Neural Information Processing Systems}, pages
  2312--2320, 2011.

\bibitem[Agarwal et~al.(2016)Agarwal, Bird, Cozowicz, Hoang, Langford, Lee, Li,
  Melamed, Oshri, Ribas, et~al.]{AgarwalEtAl16}
A.~Agarwal, S.~Bird, M.~Cozowicz, L.~Hoang, J.~Langford, S.~Lee, J.~Li,
  D.~Melamed, G.~Oshri, O.~Ribas, et~al.
\newblock Making contextual decisions with low technical debt.
\newblock \emph{arXiv preprint arXiv:1606.03966}, 2016.

\bibitem[Agrawal and Goyal(2013)]{agrawal13thompson}
S.~Agrawal and N.~Goyal.
\newblock Thompson sampling for contextual bandits with linear payoffs.
\newblock In S.~Dasgupta and D.~McAllester, editors, \emph{Proceedings of the
  30th International Conference on Machine Learning}, volume~28 of
  \emph{Proceedings of Machine Learning Research}, pages 127--135, Atlanta,
  Georgia, USA, 17--19 Jun 2013. PMLR.
\newblock URL \url{https://proceedings.mlr.press/v28/agrawal13.html}.

\bibitem[Ariel et~al.(2016)Ariel, Sutherland, Henstock, Young, Drover, Sykes,
  Megicks, and Henderson]{Ariel2016Wearing}
B.~Ariel, A.~Sutherland, D.~Henstock, J.~Young, P.~Drover, J.~Sykes,
  S.~Megicks, and R.~Henderson.
\newblock Wearing body cameras increases assaults against officers and does not
  reduce police use of force: Results from a global multi-site experiment.
\newblock \emph{European Journal of Criminology}, 13\penalty0 (6):\penalty0
  744--755, 2016.
\newblock \doi{10.1177/1477370816643734}.

\bibitem[Audibert et~al.(2010)Audibert, Bubeck, and Munos]{AudibertBuMu10}
J.-Y. Audibert, S.~Bubeck, and R.~Munos.
\newblock Best arm identification in multi-armed bandits.
\newblock In \emph{Proceedings of the Twenty Third Annual Conference on
  Computational Learning Theory}, pages 41--53, 2010.

\bibitem[Bakshy et~al.(2018{\natexlab{a}})Bakshy, Dworkin, Karrer, Kashin,
  Letham, Murthy, and Singh]{BakshyEtAl18}
E.~Bakshy, L.~Dworkin, B.~Karrer, K.~Kashin, B.~Letham, A.~Murthy, and
  S.~Singh.
\newblock Ae: A domain-agnostic platform for adaptive experimentation.
\newblock In \emph{Neural Information Processing Systems Workshop on Systems
  for Machine Learning}, pages 1--8, 2018{\natexlab{a}}.

\bibitem[Bakshy et~al.(2018{\natexlab{b}})Bakshy, Dworkin, Karrer, Kashin,
  Letham, Murthy, and Singh]{bakshy2018ae}
E.~Bakshy, L.~Dworkin, B.~Karrer, K.~Kashin, B.~Letham, A.~Murthy, and
  S.~Singh.
\newblock Ae: A domain-agnostic platform for adaptive experimentation.
\newblock In \emph{Conference on neural information processing systems}, pages
  1--8, 2018{\natexlab{b}}.

\bibitem[Bakshy et~al.(2019)]{bakshy2019ax}
E.~Bakshy et~al.
\newblock Ax: Adaptive experimentation platform.
\newblock \url{https://ax.dev}, 2019.
\newblock Accessed: 2024-06-03.

\bibitem[Balandat et~al.(2020)Balandat, Karrer, Jiang, Daulton, Letham, Wilson,
  and Bakshy]{balandat2020botorch}
M.~Balandat, B.~Karrer, D.~R. Jiang, S.~Daulton, B.~Letham, A.~G. Wilson, and
  E.~Bakshy.
\newblock Botorch: A framework for efficient monte-carlo bayesian optimization.
\newblock In \emph{Proceedings of the 34th Conference on Neural Information
  Processing Systems (NeurIPS) 2020}, Vancouver, Canada, 2020. NeurIPS.
\newblock URL \url{https://botorch.org}.

\bibitem[Bietti et~al.(2021)Bietti, Agarwal, and Langford]{bietti2021bakeoff}
A.~Bietti, A.~Agarwal, and J.~Langford.
\newblock A contextual bandit bake-off.
\newblock \emph{Journal of Machine Learning Research}, 22:\penalty0 1--49,
  2021.
\newblock Submitted 12/18; Revised 1/21; Published 5/21.

\bibitem[Bou et~al.(2023)Bou, Bettini, Dittert, Kumar, Sodhani, Yang,
  De~Fabritiis, and Moens]{bou2023torchrl}
A.~Bou, M.~Bettini, S.~Dittert, V.~Kumar, S.~Sodhani, X.~Yang, G.~De~Fabritiis,
  and V.~Moens.
\newblock Torchrl: A data-driven decision-making library for pytorch.
\newblock \emph{arXiv preprint arXiv:2306.00577}, 2023.

\bibitem[Brockman et~al.(2016)Brockman, Cheung, Pettersson, Schneider,
  Schulman, Tang, and Zaremba]{BrockmanChPeScScTaZa16}
G.~Brockman, V.~Cheung, L.~Pettersson, J.~Schneider, J.~Schulman, J.~Tang, and
  W.~Zaremba.
\newblock Open{AI} {G}ym.
\newblock \emph{arXiv: 1606.01540 [cs.LG]}, 2016.

\bibitem[Bubeck et~al.(2009)Bubeck, Munos, and Stoltz]{BubeckMuSt09}
S.~Bubeck, R.~Munos, and G.~Stoltz.
\newblock Pure exploration in multi-armed bandits problems.
\newblock In \emph{International Conference on Algorithmic Learning Theory},
  pages 23--37. Springer, 2009.

\bibitem[Carpentier and Locatelli(2016)]{CarpentierLo16}
A.~Carpentier and A.~Locatelli.
\newblock Tight (lower) bounds for the fixed budget best arm identification
  bandit problem.
\newblock In \emph{Proceedings of the Twenty Ninth Annual Conference on
  Computational Learning Theory}, pages 590--604. PMLR, 2016.

\bibitem[Che and Namkoong(2023)]{che2023adaptive}
E.~Che and H.~Namkoong.
\newblock Adaptive experimentation at scale: A computational framework for
  flexible batches.
\newblock \emph{arXiv:2303.11582 [cs.LG]}, 2023.

\bibitem[Chen et~al.(2015)Chen, Chick, Lee, and Pujowidianto]{ChenChLePu15}
C.-H. Chen, S.~E. Chick, L.~H. Lee, and N.~A. Pujowidianto.
\newblock Ranking and selection: efficient simulation budget allocation.
\newblock \emph{Handbook of Simulation Optimization}, pages 45--80, 2015.

\bibitem[Corson et~al.(1991)Corson, Decker, Dunstan, and
  Kerachsky]{corson1991pennsylvania}
W.~Corson, P.~Decker, S.~Dunstan, and S.~Kerachsky.
\newblock Pennsylvania reemployment bonus demonstration: Final report.
\newblock Technical report, Mathematica Policy Research, Inc., Princeton, NJ,
  September 1991.
\newblock Submitted to USDOL/ETA/UIS, Contract No: 99-7-0805-04-137-01, MPR No:
  7757.

\bibitem[Daulton et~al.(2020)Daulton, Balandat, and Bakshy]{daulton20diff}
S.~Daulton, M.~Balandat, and E.~Bakshy.
\newblock Differentiable expected hypervolume improvement for parallel
  multi-objective bayesian optimization.
\newblock In H.~Larochelle, M.~Ranzato, R.~Hadsell, M.~Balcan, and H.~Lin,
  editors, \emph{Advances in Neural Information Processing Systems}, volume~33,
  pages 9851--9864. Curran Associates, Inc., 2020.
\newblock URL
  \url{https://proceedings.neurips.cc/paper_files/paper/2020/file/6fec24eac8f18ed793f5eaad3dd7977c-Paper.pdf}.

\bibitem[Daulton et~al.(2021)Daulton, Balandat, and Bakshy]{daulton21par}
S.~Daulton, M.~Balandat, and E.~Bakshy.
\newblock Parallel bayesian optimization of multiple noisy objectives with
  expected hypervolume improvement.
\newblock In M.~Ranzato, A.~Beygelzimer, Y.~Dauphin, P.~Liang, and J.~W.
  Vaughan, editors, \emph{Advances in Neural Information Processing Systems},
  volume~34, pages 2187--2200. Curran Associates, Inc., 2021.
\newblock URL
  \url{https://proceedings.neurips.cc/paper_files/paper/2021/file/11704817e347269b7254e744b5e22dac-Paper.pdf}.

\bibitem[Diamantopoulos et~al.(2020)Diamantopoulos, Wong, Mattos,
  Gerostathopoulos, Wardrop, Mao, and McFarland]{diamantopoulos2020engineering}
N.~Diamantopoulos, J.~Wong, D.~I. Mattos, I.~Gerostathopoulos, M.~Wardrop,
  T.~Mao, and C.~McFarland.
\newblock Engineering for a science-centric experimentation platform.
\newblock In \emph{Proceedings of the ACM/IEEE 42nd International Conference on
  Software Engineering: Software Engineering in Practice}, pages 191--200,
  2020.

\bibitem[Egami and Hartman(2022)]{EgamiHa22}
N.~Egami and E.~Hartman.
\newblock Elements of external validity: Framework, design, and analysis.
\newblock \emph{American Political Science Review}, page 1–19, 2022.

\bibitem[Even-Dar et~al.(2006)Even-Dar, Mannor, and Mansour]{EvenDarMaMa06}
E.~Even-Dar, S.~Mannor, and Y.~Mansour.
\newblock Action elimination and stopping conditions for the multi-armed bandit
  and reinforcement learning problems.
\newblock \emph{Journal of Machine Learning Research}, 7:\penalty0 1079--1105,
  2006.

\bibitem[Fiez et~al.(2024)Fiez, Nassif, Chen, Gamez, and Jain]{fiez2024best}
T.~Fiez, H.~Nassif, Y.-C. Chen, S.~Gamez, and L.~Jain.
\newblock Best of three worlds: Adaptive experimentation for digital marketing
  in practice.
\newblock In \emph{Proceedings of the ACM on Web Conference 2024}, pages
  3586--3597, 2024.

\bibitem[Frazier et~al.(2008)Frazier, Powell, and
  Dayanik]{frazier2008knowledge}
P.~I. Frazier, W.~B. Powell, and S.~Dayanik.
\newblock A knowledge-gradient policy for sequential information collection.
\newblock \emph{SIAM Journal on Control and Optimization}, 47\penalty0
  (5):\penalty0 2410--2439, 2008.

\bibitem[Gabillon et~al.(2012)Gabillon, Ghavamzadeh, and
  Lazaric]{GabillonGhLa12}
V.~Gabillon, M.~Ghavamzadeh, and A.~Lazaric.
\newblock Best arm identification: A unified approach to fixed budget and fixed
  confidence.
\newblock In \emph{Advances in Neural Information Processing Systems 12},
  volume~25, 2012.

\bibitem[Gao et~al.(2022)Gao, Li, Lei, Chen, Li, Jiang, He, Mao, and
  Chua]{gao2022kuairec}
C.~Gao, S.~Li, W.~Lei, J.~Chen, B.~Li, P.~Jiang, X.~He, J.~Mao, and T.-S. Chua.
\newblock Kuairec: A fully-observed dataset and insights for evaluating
  recommender systems.
\newblock In \emph{Proceedings of the 31st ACM International Conference on
  Information and Knowledge Management (CIKM)}, pages 1--11, Atlanta, GA, USA,
  2022. ACM.
\newblock \doi{10.1145/3511808.3557220}.
\newblock URL \url{https://doi.org/10.1145/3511808.3557220}.

\bibitem[Garivier and Kaufmann(2016)]{GarivierKa16}
A.~Garivier and E.~Kaufmann.
\newblock Optimal best arm identification with fixed confidence.
\newblock In \emph{Conference on Learning Theory}, pages 998--1027. PMLR, 2016.

\bibitem[Glynn and Juneja(2004)]{GlynnJu04}
P.~W. Glynn and S.~Juneja.
\newblock A large deviations perspective on ordinal optimization.
\newblock In \emph{Proceedings of the 2004 Winter Simulation Conference}, pages
  577--586. IEEE, 2004.

\bibitem[Gupta et~al.(2018)Gupta, Ulanova, Bhardwaj, Dmitriev, Raff, and
  Fabijan]{gupta2018anatomy}
S.~Gupta, L.~Ulanova, S.~Bhardwaj, P.~Dmitriev, P.~Raff, and A.~Fabijan.
\newblock The anatomy of a large-scale experimentation platform.
\newblock In \emph{2018 IEEE International Conference on Software Architecture
  (ICSA)}, pages 1--109. IEEE, 2018.

\bibitem[Gupta et~al.(2019)Gupta, Kohavi, Tang, Xu, Andersen, Bakshy, Cardin,
  Chandran, Chen, Coey, et~al.]{gupta2019top}
S.~Gupta, R.~Kohavi, D.~Tang, Y.~Xu, R.~Andersen, E.~Bakshy, N.~Cardin,
  S.~Chandran, N.~Chen, D.~Coey, et~al.
\newblock Top challenges from the first practical online controlled experiments
  summit.
\newblock \emph{ACM SIGKDD Explorations Newsletter}, 21\penalty0 (1):\penalty0
  20--35, 2019.

\bibitem[Hamilton et~al.(2001)Hamilton, Freedman, Gennetian, Michalopoulos,
  Walter, Adams-Ciardullo, Gassman-Pines, McGroder, Zaslow, Brooks, Ahluwalia,
  Small, and Ricchetti]{Hamilton2001National}
G.~Hamilton, S.~Freedman, L.~Gennetian, C.~Michalopoulos, J.~Walter,
  D.~Adams-Ciardullo, A.~Gassman-Pines, S.~McGroder, M.~Zaslow, J.~Brooks,
  S.~Ahluwalia, E.~Small, and B.~Ricchetti.
\newblock \emph{National Evaluation of Welfare-to-Work Strategies: How
  Effective Are Different Welfare-to-Work Approaches? Five-Year Adult and Child
  Impacts for Eleven Programs}.
\newblock Manpower Demonstration Research Corporation and Child Trends, U.S.,
  December 2001.

\bibitem[Hong et~al.(2015)Hong, Nelson, and Xu]{HongNeXu15}
L.~J. Hong, B.~L. Nelson, and J.~Xu.
\newblock Discrete optimization via simulation.
\newblock In \emph{Handbook of Simulation Optimization}, pages 9--44. Springer,
  2015.

\bibitem[Huang et~al.(2022)Huang, Dossa, Ye, Braga, Chakraborty, Mehta, and
  Ara{\~A}{\v{s}}jo]{huang2022cleanrl}
S.~Huang, R.~F.~J. Dossa, C.~Ye, J.~Braga, D.~Chakraborty, K.~Mehta, and J.~G.
  Ara{\~A}{\v{s}}jo.
\newblock Cleanrl: High-quality single-file implementations of deep
  reinforcement learning algorithms.
\newblock \emph{Journal of Machine Learning Research}, 23\penalty0
  (274):\penalty0 1--18, 2022.

\bibitem[Ioannidis et~al.(2017)Ioannidis, Stanley, and
  Doucouliagos]{ioannidis2017power}
J.~P.~A. Ioannidis, T.~D. Stanley, and H.~Doucouliagos.
\newblock The power of bias in economics research.
\newblock \emph{The Economic Journal}, 127:\penalty0 F236--F265, 2017.
\newblock \doi{10.1111/ecoj.12461}.
\newblock URL \url{https://doi.org/10.1111/ecoj.12461}.

\bibitem[Jamieson and Nowak(2014)]{JamiesonNo14}
K.~Jamieson and R.~Nowak.
\newblock Best-arm identification algorithms for multi-armed bandits in the
  fixed confidence setting.
\newblock In \emph{2014 48th Annual Conference on Information Sciences and
  Systems (CISS)}, pages 1--6. IEEE, 2014.

\bibitem[Jamieson et~al.(2014)Jamieson, Malloy, Nowak, and
  Bubeck]{JamiesonMaNoBu14}
K.~Jamieson, M.~Malloy, R.~Nowak, and S.~Bubeck.
\newblock lil ucb : An optimal exploration algorithm for multi-armed bandits.
\newblock In \emph{Proceedings of the Twenty Seventh Annual Conference on
  Computational Learning Theory}, pages 423--439. PMLR, 2014.

\bibitem[Jeong and Namkoong(2020)]{JeongNa22}
S.~Jeong and H.~Namkoong.
\newblock Assessing external validity over worst-case subpopulations.
\newblock \emph{arXiv:2007.02411 [stat.ML]}, 2020.

\bibitem[Jiang et~al.(2020)Jiang, Jiang, Balandat, Karrer, Gardner, and
  Garnett]{jiang2020efficient}
S.~Jiang, D.~Jiang, M.~Balandat, B.~Karrer, J.~Gardner, and R.~Garnett.
\newblock Efficient nonmyopic bayesian optimization via one-shot multi-step
  trees.
\newblock \emph{Advances in Neural Information Processing Systems},
  33:\penalty0 18039--18049, 2020.

\bibitem[Jones et~al.(1998)Jones, Schonlau, and Welch]{jones1998efficient}
D.~R. Jones, M.~Schonlau, and W.~J. Welch.
\newblock Efficient global optimization of expensive black-box functions.
\newblock \emph{Journal of Global optimization}, 13:\penalty0 455--492, 1998.

\bibitem[Kandasamy et~al.(2020)Kandasamy, Vysyaraju, Neiswanger, Paria,
  Collins, Schneider, Poczos, and Xing]{dragonfly2020}
K.~Kandasamy, K.~R. Vysyaraju, W.~Neiswanger, B.~Paria, C.~R. Collins,
  J.~Schneider, B.~Poczos, and E.~P. Xing.
\newblock Tuning hyperparameters without grad students: Scalable and robust
  bayesian optimisation with dragonfly.
\newblock \emph{Journal of Machine Learning Research}, 21\penalty0
  (81):\penalty0 1--27, 2020.
\newblock URL \url{http://jmlr.org/papers/v21/18-223.html}.

\bibitem[Karnin et~al.(2013)Karnin, Koren, and Somekh]{KarninKoSo13}
Z.~Karnin, T.~Koren, and O.~Somekh.
\newblock Almost optimal exploration in multi-armed bandits.
\newblock In \emph{Proceedings of the 30th International Conference on Machine
  Learning}, pages 1238--1246, 2013.

\bibitem[Kaufmann and Kalyanakrishnan(2013)]{KaufmannKa13}
E.~Kaufmann and S.~Kalyanakrishnan.
\newblock Information complexity in bandit subset selection.
\newblock In \emph{Proceedings of the Twenty Sixth Annual Conference on
  Computational Learning Theory}, pages 228--251. PMLR, 2013.

\bibitem[Kaufmann et~al.(2016)Kaufmann, Capp{\'e}, and
  Garivier]{KaufmannCaGa16}
E.~Kaufmann, O.~Capp{\'e}, and A.~Garivier.
\newblock On the complexity of best-arm identification in multi-armed bandit
  models.
\newblock \emph{Journal of Machine Learning Research}, 17\penalty0
  (1):\penalty0 1--42, 2016.

\bibitem[Kemple et~al.(1995)Kemple, Friedlander, and
  Fellerath]{Kemple1995Florida}
J.~J. Kemple, D.~Friedlander, and V.~Fellerath.
\newblock \emph{Florida's Project Independence: Benefits, Costs, and Two-Year
  Impacts of Florida's JOBS Program}.
\newblock U.S., April 1995.

\bibitem[Kohavi et~al.(2009)Kohavi, Longbotham, Sommerfield, and
  Henne]{KohaviLoSoHe09}
R.~Kohavi, R.~Longbotham, D.~Sommerfield, and R.~M. Henne.
\newblock Controlled experiments on the web: survey and practical guide.
\newblock \emph{Data Mining and Knowledge Discovery}, 18\penalty0 (1):\penalty0
  140--181, 2009.

\bibitem[Kohavi et~al.(2020)Kohavi, Tang, and Xu]{kohavi2020trustworthy}
R.~Kohavi, D.~Tang, and Y.~Xu.
\newblock \emph{Trustworthy online controlled experiments: A practical guide to
  a/b testing}.
\newblock Cambridge University Press, 2020.

\bibitem[Lam et~al.(2016)Lam, Willcox, and Wolpert]{lam2016bayesian}
R.~Lam, K.~Willcox, and D.~H. Wolpert.
\newblock Bayesian optimization with a finite budget: An approximate dynamic
  programming approach.
\newblock \emph{Advances in Neural Information Processing Systems}, 29, 2016.

\bibitem[Langford et~al.(2012)Langford, Li, and Strehl]{vowpal2012}
J.~Langford, L.~Li, and A.~Strehl.
\newblock Vowpal wabbit, 2012.
\newblock URL \url{https://github.com/VowpalWabbit/vowpal_wabbit}.

\bibitem[Lattimore and Szepesv{\'a}ri(2019)]{LattimoreSz19}
T.~Lattimore and C.~Szepesv{\'a}ri.
\newblock \emph{Bandit algorithms}.
\newblock Cambridge, 2019.

\bibitem[Lefortier et~al.(2016)Lefortier, Swaminathan, Gu, Joachims, and
  de~Rijke]{lefortier2016large}
D.~Lefortier, A.~Swaminathan, X.~Gu, T.~Joachims, and M.~de~Rijke.
\newblock Large-scale validation of counterfactual learning methods: A
  test-bed.
\newblock \emph{arXiv preprint arXiv:1612.00367}, 2016.

\bibitem[Li et~al.(2010)Li, Chu, Langford, and Schapire]{li2010contextual}
L.~Li, W.~Chu, J.~Langford, and R.~E. Schapire.
\newblock A contextual-bandit approach to personalized news article
  recommendation.
\newblock In \emph{Proceedings of the International World Wide Web Conference
  (WWW)}, Raleigh, North Carolina, USA, 2010. WWW.
\newblock A version of this paper appears at WWW 2010, April 26-30, 2010,
  Raleigh, North Carolina, USA.

\bibitem[Liang et~al.(2018)Liang, Liaw, Nishihara, Moritz, Fox, Goldberg,
  Gonzalez, Jordan, and Stoica]{liang2018rllib}
E.~Liang, R.~Liaw, R.~Nishihara, P.~Moritz, R.~Fox, K.~Goldberg, J.~Gonzalez,
  M.~Jordan, and I.~Stoica.
\newblock Rllib: Abstractions for distributed reinforcement learning.
\newblock In \emph{International conference on machine learning}, pages
  3053--3062. PMLR, 2018.

\bibitem[Liu et~al.(2021)Liu, Cardoso, Couturier, and McCoy]{liu2021datasets}
C.~Liu, {\^A}.~Cardoso, P.~Couturier, and E.~J. McCoy.
\newblock Datasets for online controlled experiments.
\newblock \emph{arXiv preprint arXiv:2111.10198}, 2021.

\bibitem[Mannor and Tsitsiklis(2004)]{MannorTs04}
S.~Mannor and J.~N. Tsitsiklis.
\newblock The sample complexity of exploration in the multi-armed bandit
  problem.
\newblock \emph{Journal of Machine Learning Research}, 5\penalty0
  (Jun):\penalty0 623--648, 2004.

\bibitem[Meager(2019)]{Meager2019Understanding}
R.~Meager.
\newblock Understanding the average impact of microcredit expansions: A
  bayesian hierarchical analysis of seven randomized experiments.
\newblock \emph{American Economic Journal: Applied Economics}, 11\penalty0
  (1):\penalty0 57--91, 2019.

\bibitem[Min et~al.(2019)Min, Maglaras, and Moallemi]{min2019thompson}
S.~Min, C.~Maglaras, and C.~C. Moallemi.
\newblock Thompson sampling with information relaxation penalties.
\newblock \emph{Advances in neural information processing systems}, 32, 2019.

\bibitem[Min et~al.(2020)Min, Moallemi, and Russo]{min2020policy}
S.~Min, C.~C. Moallemi, and D.~J. Russo.
\newblock Policy gradient optimization of thompson sampling policies.
\newblock \emph{arXiv preprint arXiv:2006.16507}, 2020.

\bibitem[Namkoong et~al.(2020)Namkoong, Daulton, and Bakshy]{NamkoongDaBa20}
H.~Namkoong, S.~Daulton, and E.~Bakshy.
\newblock Distilled thompson sampling: Practical and efficient thompson
  sampling via imitation learning.
\newblock \emph{arXiv:2011.14266 [cs.LG]}, 2020.

\bibitem[{National Center for Health Statistics}(2019)]{NHIS2010-2019}
{National Center for Health Statistics}.
\newblock {National Health Interview Survey, 2010-2019}.
\newblock \url{https://www.cdc.gov/nchs/nhis/index.htm}, 2019.
\newblock Accessed: 2024-06-05.

\bibitem[Offer-Westort et~al.(2020)Offer-Westort, Coppock, and
  Green]{OfferWestortCoGr20}
M.~Offer-Westort, A.~Coppock, and D.~P. Green.
\newblock Adaptive experimental design: Prospects and applications in political
  science.
\newblock \emph{SSRN 3364402}, 2020.
\newblock URL \url{http://dx.doi.org/10.2139/ssrn.3364402}.

\bibitem[Owen et~al.(2021)Owen, Browder, Letham, Stocek, Tymms, and
  Shvartsman]{owen21psycho}
L.~Owen, J.~Browder, B.~Letham, G.~Stocek, C.~Tymms, and M.~Shvartsman.
\newblock Adaptive nonparametric psychophysics, 2021.
\newblock URL \url{https://arxiv.org/abs/2104.09549}.

\bibitem[Qin and Russo(2023)]{QinRusso2023}
C.~Qin and D.~Russo.
\newblock Adaptive experimentation in the presence of exogenous nonstationary
  variation.
\newblock \emph{arXiv:2202.09036 [cs.LG]}, 2023.

\bibitem[Qin et~al.(2017)Qin, Klabjan, and Russo]{QinKlRu17}
C.~Qin, D.~Klabjan, and D.~Russo.
\newblock Improving the expected improvement algorithm.
\newblock In \emph{Advances in Neural Information Processing Systems 17}, 2017.

\bibitem[Raffin et~al.(2021)Raffin, Hill, Gleave, Kanervisto, Ernestus, and
  Dormann]{raffin2021stable}
A.~Raffin, A.~Hill, A.~Gleave, A.~Kanervisto, M.~Ernestus, and N.~Dormann.
\newblock Stable-baselines3: Reliable reinforcement learning implementations.
\newblock \emph{Journal of Machine Learning Research}, 22\penalty0
  (268):\penalty0 1--8, 2021.

\bibitem[Riccio and Friedlander(1992)]{Riccio1992GAIN}
J.~Riccio and D.~Friedlander.
\newblock \emph{GAIN: Program Strategies, Participation Patterns, and
  First-Year Impacts in Six Counties}.
\newblock Manpower Demonstration Research Corporation, California, 1992.
\newblock California's Greater Avenues for Independence Program.

\bibitem[Riquelme et~al.(2018)Riquelme, Tucker, and Snoek]{riquelme2018deep}
C.~Riquelme, G.~Tucker, and J.~Snoek.
\newblock Deep bayesian bandits showdown: An empirical comparison of bayesian
  deep networks for thompson sampling.
\newblock In \emph{Proceedings of the International Conference on Learning
  Representations (ICLR)}, Google Brain, 2018. ICLR, ICLR.
\newblock Published as a conference paper at ICLR 2018.

\bibitem[Rohde et~al.(2018)Rohde, Bonner, Dunlop, Vasile, and
  Karatzoglou]{rohde2018recogym}
D.~Rohde, S.~Bonner, T.~Dunlop, F.~Vasile, and A.~Karatzoglou.
\newblock Recogym: A reinforcement learning environment for the problem of
  product recommendation in online advertising.
\newblock In \emph{Proceedings of the 12th ACM Conference on Recommender
  Systems (RecSys'18)}, page~5, New York, NY, USA, 2018. ACM.
\newblock URL \url{https://arxiv.org/abs/1808.00720}.
\newblock 14 Sep 2018.

\bibitem[Russo(2020)]{Russo20}
D.~Russo.
\newblock Simple bayesian algorithms for best-arm identification.
\newblock \emph{Operations Research}, 68\penalty0 (6):\penalty0 1625--1647,
  2020.

\bibitem[Saito et~al.(2021)Saito, Aihara, Matsutani, and Narita]{saito2021open}
Y.~Saito, S.~Aihara, M.~Matsutani, and Y.~Narita.
\newblock Open bandit dataset and pipeline: Towards realistic and reproducible
  off-policy evaluation.
\newblock In \emph{Proceedings of the 35th Conference on Neural Information
  Processing Systems (NeurIPS) Track on Datasets and Benchmarks}, Tokyo, Japan,
  2021. NeurIPS.
\newblock This work was done when YS was at Hanjuku-kaso Co Ltd, Tokyo, Japan.

\bibitem[Sculley et~al.(2015)Sculley, Holt, Golovin, Davydov, Phillips, Ebner,
  Chaudhary, Young, Crespo, and Dennison]{SculleyEtAl15}
D.~Sculley, G.~Holt, D.~Golovin, E.~Davydov, T.~Phillips, D.~Ebner,
  V.~Chaudhary, M.~Young, J.-F. Crespo, and D.~Dennison.
\newblock Hidden technical debt in machine learning systems.
\newblock In \emph{Advances in Neural Information Processing Systems 28}, pages
  2503--2511, 2015.

\bibitem[Wahlstrom et~al.(2014)Wahlstrom, Dretzke, Gordon, Peterson, Edwards,
  and Gdula]{wahlstrom2014examining}
K.~Wahlstrom, B.~Dretzke, M.~Gordon, K.~Peterson, K.~Edwards, and J.~Gdula.
\newblock Examining the impact of later high school start times on the health
  and academic performance of high school students: A multi-site study.
\newblock 2014.

\bibitem[Weiss et~al.(2017)Weiss, Bloom, Savitz, and Gupta]{weiss2017effects}
M.~J. Weiss, H.~S. Bloom, N.~V. Savitz, and H.~Gupta.
\newblock How much do the effects of education and training programs vary
  across sites? evidence from past multisite randomized trials.
\newblock \emph{Journal of Research on Educational Effectiveness}, 10\penalty0
  (6), 2017.
\newblock \doi{10.1080/19345747.2017.1300719}.

\bibitem[Weng et~al.(2022)Weng, Chen, Yan, You, Duburcq, Zhang, Su, Su, and
  Zhu]{weng2022tianshou}
J.~Weng, H.~Chen, D.~Yan, K.~You, A.~Duburcq, M.~Zhang, Y.~Su, H.~Su, and
  J.~Zhu.
\newblock Tianshou: A highly modularized deep reinforcement learning library.
\newblock \emph{Journal of Machine Learning Research}, 23\penalty0
  (267):\penalty0 1--6, 2022.

\bibitem[Wu et~al.(2020)Wu, Qiao, Chen, Wu, Qi, Lian, Liu, Xie, Gao, Wu, and
  Zhou]{wu2020mind}
F.~Wu, Y.~Qiao, J.-H. Chen, C.~Wu, T.~Qi, J.~Lian, D.~Liu, X.~Xie, J.~Gao,
  W.~Wu, and M.~Zhou.
\newblock Mind: A large-scale dataset for news recommendation.
\newblock In \emph{Proceedings of the Annual Meeting of the Association for
  Computational Linguistics (ACL)}. ACL, 2020.

\bibitem[Xia et~al.(2015)Xia, Li, Qin, Yu, and Liu]{xia2015thompson}
Y.~Xia, H.~Li, T.~Qin, N.~Yu, and T.-Y. Liu.
\newblock Thompson sampling for budgeted multi-armed bandits, 2015.

\bibitem[Xu et~al.(2015)Xu, Chen, Fernandez, Sinno, and
  Bhasin]{xu2015infrastructure}
Y.~Xu, N.~Chen, A.~Fernandez, O.~Sinno, and A.~Bhasin.
\newblock From infrastructure to culture: A/b testing challenges in large scale
  social networks.
\newblock In \emph{Proceedings of the 21th ACM SIGKDD international conference
  on knowledge discovery and data mining}, pages 2227--2236, 2015.

\bibitem[Zhang et~al.(2020)Zhang, Janson, and Murphy]{ZhangJaMu20}
K.~Zhang, L.~Janson, and S.~Murphy.
\newblock Inference for batched bandits.
\newblock \emph{Advances in Neural Information Processing Systems 20},
  33:\penalty0 9818--9829, 2020.

\end{thebibliography}
}

\appendix

\end{document}